\documentclass[12pt]{elsarticle}
\usepackage{amsmath,amsfonts}
\usepackage{algorithmic}
\usepackage{array}
\usepackage{textcomp}
\usepackage{stfloats}
\usepackage{url}
\usepackage{verbatim}
\usepackage{graphicx}
\usepackage{subcaption}
\usepackage{bm}
\usepackage{float} 
\usepackage{balance}
\usepackage{algorithm}
\usepackage{natbib}
\setcitestyle{numbers,square}

\makeatletter

\usepackage{hyperref}

\usepackage{color}
\usepackage{booktabs}
\usepackage{array, caption, threeparttable}
\usepackage[font=small,labelfont=bf,labelsep=none]{caption}
\usepackage{stfloats}
\usepackage[figuresleft]{rotating}

\journal{}
\makeatother

\begin{document}

\begin{frontmatter}

\title{A Comprehensive Study for Edge Detection: Ideal-Prior Guidance Learning, High-Precision Architecture, and Better Distinguishability Evaluation Benchmark}

\author{Hao Shu\corref{cor1}}
\ead{Hao_B_Shu@163.com}
\cortext[cor1]{}

\affiliation{organization={Shenzhen University},
	city={Shenzhen},
	postcode={518060}, 
	state={Guangdong},
	country={P. R. China}}
\affiliation{organization={Sun-Yat-Sen University},
		city={Shenzhen},
		postcode={518060}, 
		state={Guangdong},
		country={P. R. China}}

\begin{abstract}
Image edge detection (ED) requires specialized architectures, reliable supervision, and rigorous evaluation criteria to ensure accurate localization. In this work, we present a framework for high-precision ED that jointly addresses architectural design, data supervision, and evaluation consistency. We propose SDPED, a compact ED model built upon Cascaded Skipping Density Blocks (CSDB), motivated by a task-adaptive architectural transfer from image super-resolution. By re-engineering texture-oriented structures for ED, SDPED effectively differentiates textures from edges while preserving fine spatial precision. Extensive experiments on four benchmark datasets (BRIND, UDED, MDBD, and BIPED2) demonstrate consistent performance improvements, particularly in Average Precision (AP), with gains of up to 22.5\% on MDBD and 11.8\% on BIPED2. In addition, we introduce an ideal-prior guidance strategy that incorporates noiseless data into training by treating labels as noise-free samples, providing a practical means to mitigate the subjectivity and noise inherent in human annotations. To enable better distinguishability and resolution-independent evaluation, we further study the criterion for assessing ED models, which suggests a stricter 1-pixel error tolerance distance. Overall, this work offers a coherent solution for high-precision ED and provides insights applicable to precision-oriented modeling in low-level and soft-computing-based vision tasks. Codes can be found on https://github.com/Hao-B-Shu/SDPED.

\end{abstract}


\begin{highlights}
	\item A compact edge detection architecture based on cascaded skipping density blocks.
	\item Task-adaptive architectural transfer from super-resolution to edge detection.
	\item An ideal-prior guidance strategy leveraging noiseless annotations to mitigate label noise.
	\item Resolution-independent and stricter evaluation for consistent and better distinguishability model assessment.
	\item Consistent precision improvements validated across multiple datasets and partitions.
\end{highlights}


\end{frontmatter}

\section{Introduction}

Image edge detection (ED) is a fundamental task in computer vision and has been studied for decades, with extensive applications in high-level tasks such as object detection\cite{ZD2007An} and image segmentation\cite{MR2011Edge}. Early ED methods, such as the Sobel operator\cite{K1983On} and the Canny algorithm\cite{C1986A}, relied on gradient-based features. Later, more sophisticated statistical approaches were introduced\cite{KY2003Statistical,AM2011Contour}. In recent years, machine learning has become indispensable in nearly all computer vision tasks, leading to the emergence of numerous learning-based ED algorithms\cite{MF2004Learning,DT2006Supervised,R2008Multi,LZ2013Sketch,GL2014N,DZ2015Fast}. Currently, convolutional neural network (CNN)-based models achieve state-of-the-art (SOTA) performance in ED tasks \cite{SW2015Deepcontour,XT2015Holistically,HZ2019Bi,LK2021Revisiting,SL2021Pixel,HX2022Unmixing,SS2023Dense}, while recently, transformer-based approaches have also been explored\cite{PH2022Edter}.

The goal of ED is to identify edges in an image, which typically correspond to pixels with significant changes in intensity compared with their neighbors. However, textural areas can also contain rapid intensity changes that do not correspond to edges. Classical ED algorithms such as Canny\cite{C1986A}, despite their efficiency, often fail to differentiate between edges and textures due to their inability to adapt to varying data distributions. Consequently, learning-based models, particularly CNNs, have become the dominant approach. Over the past decade, CNN-based ED models have demonstrated remarkable progress, with models such as HED\cite{XT2015Holistically}, RCF\cite{LC2017Richer}, RINDNet\cite{PH2021RINDNet}, BDCN\cite{HZ2022BDCN}, DexiNed\cite{SS2023Dense}, and even lightweight models such as TEED\cite{SL2023Tiny} achieving $F$-scores comparable to or exceeding human performance under commonly adopted evaluation procedures.

\begin{figure}
\centering
\includegraphics[width=5.5in]{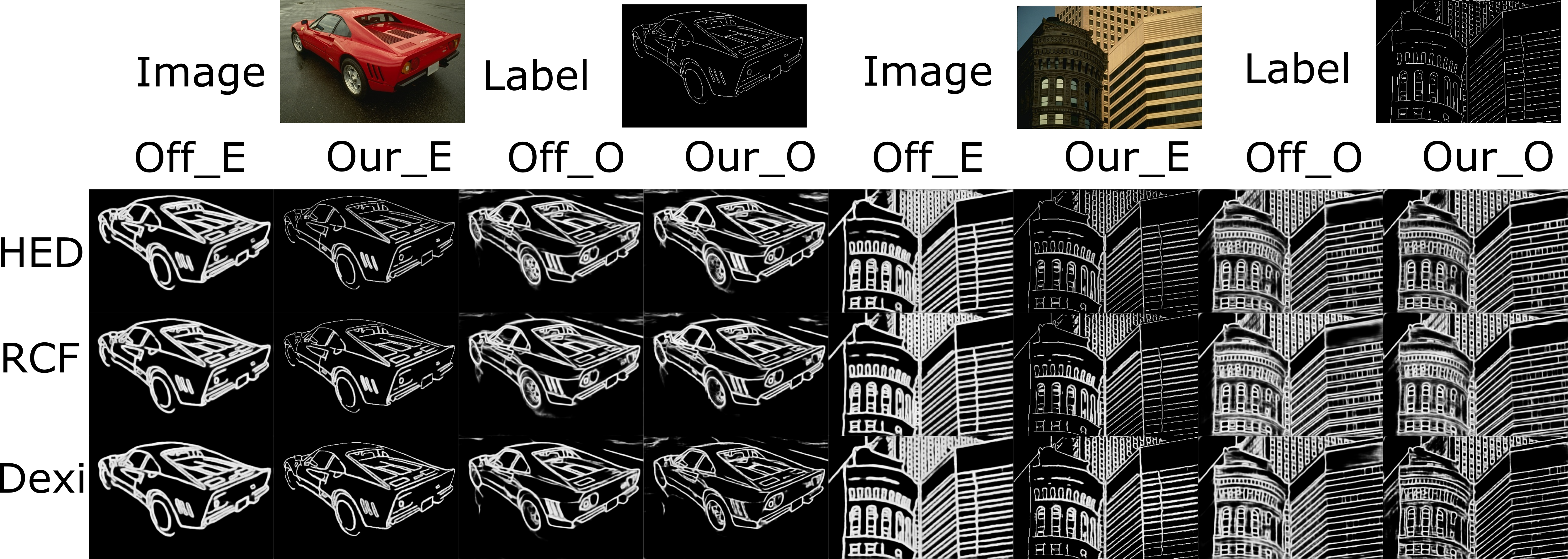}
\caption{\quad Predictions of three previous ED models on the BRIND dataset. \textbf{\textit{Off$\_$E}} denotes models retrained using the official data augmentation and tested on edge maps, while \textbf{\textit{Our$\_$E}} represents models retrained using our proposed data augmentation (detailed in Section \ref{sec:Noiseless}) and tested on edge maps. Similarly, \textbf{\textit{Off$\_$O}} and \textbf{\textit{Our$\_$O}} correspond to models trained with respective augmentation strategies but tested on standard images. Our augmentation method produces sharper predictions on edge maps and potentially enhances performance on standard images.}
\label{fig:Noiseless}
\end{figure}

Despite these advancements, several challenges remain in current ED research. In this paper, we focus on the following two key issues:

\begin{enumerate}
   
    \item \textbf{Suboptimal Reliability on Noiseless Inputs}: As shown in Fig \ref{fig:Noiseless}, existing models often produce suboptimal edge maps even when given noiseless input data, namely ground-truth edge maps themselves, which represent deterministic and noise-free ideal priors. Ideally, an ED model should yield nearly perfect predictions on noiseless inputs since they represent the "ideal prior" of edges. Yet, most models generate results on noiseless inputs that are similar to those on textured images, suffering from blurred or discontinuous predictions. This limitation fundamentally stems from training solely on human-annotated data, where subjectivity and annotation noise are unavoidable.
   
   \item \textbf{Insufficient Precision under Strict Constraints}: Current ED models achieve high scores only when evaluated with a large error tolerance. For example, the standard error tolerance radius is approximately 4.3 pixels for BSDS500\cite{AM2011Contour}, 11.1 pixels for BIPED\cite{SR2020Dense}, and even 16.3 pixels for MDBD\cite{MK2016A}. However, when stricter evaluation criteria are applied to ensure high-precision localization, the performance of these models drops significantly. This indicates that current architectures may lack sufficient spatial fidelity to distinguish fine edges from surrounding textures, suggesting the need for architectures explicitly designed for high-precision localization.

    \item \textbf{Optimal Error Tolerance Distance for Model Comparison}: The standard evaluation algorithm used in ED tasks is published in \cite{MF2004Learning}, which sets 0.0075 (ratio of the diagonal) as the default error tolerance distance. The default setting is inherited by most later works, regardless of the dataset change. However, whether this setting is optimal is rarely discussed. Indeed, this setting seems not so reasonable when the resolution of test images is varied. For example, if four ordinary images are spliced to one image with double resolution diagonally, then the error tolerance distance by the default setting will be doubled, but the content of the images is not changed. Therefore, high-resolution images might always obtain better scores even when they are merely spliced with lower-resolution ones. Additionally, whether the default error tolerance distance is optimal for a fixed dataset with fixed resolution is also a problem without sufficiently studied.
\end{enumerate}

To address these challenges, we propose a unified high-precision ED framework. We observe a fundamental synergy between image super-resolution (SR) and ED: both tasks rely on carefully handling textures, albeit with different task objectives. Drawing inspiration from this, we perform a task-adaptive architectural transfer by introducing Cascaded Skipping Density Blocks (CSDB). While density blocks are adept at handling textures in SR\cite{WX2021Real}, our investigation shows that their direct application to ED is insufficient for achieving precise localization. However, by re-engineering them into a cascaded skipping structure with task-aware modifications, the architecture can be effectively transferred to meet the stringent precision requirements of ED. Furthermore, we introduce a novel data augmentation strategy that integrates noiseless annotations as ideal prior guidance, which allows the model to learn from deterministic, noise-free samples as ideal guidance, thereby alleviating the performance ceiling imposed by subjective human annotations. To ensure a scientifically rigorous assessment of these improvements, we also advocate for and analyze the necessity of employing stricter evaluation standards to measure true localization precision.

The key contributions of this work are as follows:

\begin{itemize} 

\item \textbf{Ideal Prior Guidance via Noiseless Data:} We propose a novel data augmentation method that treats training labels as noise-free input samples. As far as we know, this provides the first feasible way to introduce "perfect knowledge" into ED training, which aims to mitigate the impact of human-annotation noise. This paradigm of label-to-data learning is potentially generalizable to a broad range of supervised tasks in image processing and related domains, especially in tasks where ground-truth labels can only be obtained by human annotations, which unavoidably induces noise.

\item \textbf{Task-Adaptive High-Precision Architecture:} We introduce Skipping Density Precise Edge Detection (SDPED), a parameter-efficient yet powerful model for high-performance edge detection. By incorporating CSDB—a structure not previously adapted to ED in a task-aware manner—we demonstrate that dense architectural mechanisms from SR can be transferred to ED tasks with several modifications. Notably, our models do not require multi-scale supervision, challenging the widely held belief that such supervision is essential for extracting complex edge features.

\item \textbf{Better distinguishability evaluation setting:} We rethink the fairness and effectiveness of the error tolerance used in assessing ED models, showing that a stricter error tolerance criterion allows better distinguishability across models, and pixel units should be used instead of the diagonal ratio. Based on these, a 1-pixel error tolerance distance is suggested when comparing ED models, as it is demonstrated to be optimal in some sense.

\item \textbf{Comprehensive Validation and Scientific Evaluation:} Through extensive experiments on multiple models and datasets, we demonstrate the effectiveness of our methodology.
\end{itemize}

\section{Review of Previous Work}

This section reviews previous works in the field of ED, including datasets, pre-processing techniques, ED models, loss functions, and evaluation metrics.

\subsection{Datasets, Label Refinement, and Data Augmentation}

Early ED datasets were primarily derived from segmentation tasks, such as BSDS300 and its later extension, BSDS500\cite{MF2001A}, which remains one of the most widely used benchmarks. BSDS500 consists of 500 RGB images annotated by multiple users and is frequently utilized for both segmentation and boundary detection tasks. Another well-known dataset is NYUD/NYUD2\cite{SH2012Indoor}, which contains hundreds of thousands of RGB-D images, originally designed for indoor scene analysis. Among these, 1,449 images are annotated with segmentation categories. Other commonly used datasets include PASCAL VOC\cite{EV2010The}, Microsoft COCO\cite{LM2014Microsoft}, SceneParse150\cite{ZZ2017Scene}, and Cityscape\cite{CO2016The}. However, since these datasets were not explicitly designed for ED, their labels often lack the spatial precision required for accurate edge localization.

With the increasing focus on ED tasks, several dedicated datasets have been proposed. In 2016, the MDBD dataset\cite{MK2016A} was introduced, containing 100 high-resolution (1280$\times$720) images, each annotated by multiple annotators. While initially designed for boundary detection, it has been widely adopted in ED research. In 2020, the BIPED dataset\cite{SR2020Dense} was specifically developed for ED tasks, featuring 250 high-resolution (1280$\times$720) RGB images with precise edge annotations, later refined in its BIPED2 version. In 2021, BRIND\cite{PH2021RINDNet} was introduced as a re-annotation of BSDS500, categorizing edges into four types: reflectance, illumination, normal, and depth edges. More recently, UDED\cite{SL2023Tiny}, proposed in 2023, consists of 29 high-quality images with carefully labeled edges, curated from multiple datasets.

A fundamental and persistent challenge across all ED datasets is annotation noise. Since edge labels are based on human annotations, perfect accuracy is unattainable, and inconsistencies inevitably arise both across different annotators and even within the same annotator across multiple attempts. As a result, various studies have focused on refining human-labeled data or enhancing model robustness to noisy annotations\cite{FG2023Practical, WD2024One}.

Regardless of the dataset used, data augmentation is widely recognized as essential in ED tasks. Common augmentation techniques include rotation, flipping, cropping, resizing, and gamma transformations. However, augmentation strategies vary across studies. In HED\cite{XT2015Holistically}, images are rotated at 16 different angles, flipped, and cropped to obtain 32 augmented images per sample. In RIND\cite{PH2021RINDNet}, only four rotations with flips (without cropping) are applied, resulting in eight augmented images per sample. In contrast, Dexi\cite{SS2023Dense} employs a more aggressive augmentation strategy, incorporating 16 rotations, flips, cropping, and three gamma transformations, generating 288 augmented images per sample. While extensive augmentation increases data diversity, its actual impact on edge localization performance remains an open question, and excessive augmentation can significantly increase training costs. Moreover, existing augmentation strategies operate exclusively on human-annotated data and inherently propagate annotation noise. To the best of our knowledge, no existing augmentation strategy incorporates noiseless training data to guide the model toward learning the accurate edge.

\subsection{ED Models and Loss Functions}

While recent studies have begun exploring transformer-based or mixed models\cite{PH2022Edter,JG2024EdgeNAT,WL2025Enhanced,S2025Boost}, CNN-based architectures continue to dominate the ED field. The HED model\cite{XT2015Holistically} introduced a multi-scale and multi-level output strategy, where predictions at different levels are fused, and the loss is computed as the average of all supervised outputs. RCF\cite{LC2017Richer} extends this approach by integrating multi-scale outputs with feature fusion across different layers, generating side outputs to enhance edge representation. BDCN\cite{HZ2022BDCN} applies layer-specific supervision, further improving ED accuracy. PiDiNet\cite{SL2021Pixel} introduces pixel-difference convolutions, while RIND\cite{PH2021RINDNet} predicts four edge types through a three-stage processing pipeline. Dexi\cite{SS2023Dense} leverages extremely dense skip connections to enhance feature propagation.

Most ED models struggle to produce crisp and well-localized edges without additional post-processing techniques such as non-maximal suppression (NMS). This issue may arise from the use of down-sampling operations such as max-pooling, as well as the widespread use of weighted binary cross-entropy (WBCE) as the loss function. However, techniques such as NMS are non-continuous operations and thus cannot be integrated into training pipelines, reducing the value of ED as a reliable and differentiable support cue for high-level vision tasks. Various strategies have been proposed to address this challenge. For instance, \cite{WZ2017Deep} suggests improving edge sharpness by aggregating features at larger scales during up-sampling, while \cite{DS2018Learning} explores the effectiveness of Dice loss in ED tasks. More recently, tracing loss\cite{HX2022Unmixing} has emerged as an alternative and has been adopted in TEED\cite{SL2023Tiny}, rank loss\cite{CK2024RankED} decides edges based on ordering pixels, and SWBCE\cite{S2026Symmetrization} loss models human asymmetric recognition on edge and not-edge. However, these solutions often introduce trade-offs, requiring careful model-specific or dataset-specific hyperparameter tuning, which limits their robustness and general applicability.

\subsection{Evaluation Metrics for ED}

Although a variety of evaluation metrics have been explored for ED tasks, such as root mean square error (RMSE), structural similarity index measure (SSIM), and edge ratio (ER)\cite{S2026Symmetrization}, the most-used evaluation framework follows the three key metrics: Optimal Dataset Score (ODS), Optimal Image Score (OIS), and Average Precision (AP). These metrics are derived from the $F_{\beta}$-score, defined as:

\begin{equation}
F_{\beta} = \frac{(1+\beta^2) \times \text{Precision} \times \text{Recall}}{\beta^2 \times \text{Precision} + \text{Recall}},
\end{equation}
where \textit{Precision} is calculated as:
\begin{equation}
\text{Precision} = \frac{TP}{TP+FP},
\end{equation}
and \textit{Recall} is given by:
\begin{equation}
\text{Recall} = \frac{TP}{TP+FN}.
\end{equation}
Here, \textit{TP} (true positives) refers to correctly predicted edge pixels, \textit{FP} (false positives) refers to non-edge pixels incorrectly classified as edges, and \textit{FN} (false negatives) refers to missed edge pixels. The parameter $\beta$, typically set to 1, balances precision and recall.

Most ED models utilize the evaluation algorithm from\cite{MF2004Learning}, which includes a hyperparameter, \textit{maxDist}, representing the maximum allowable error tolerance distance. The error tolerance distance is expressed as a fraction of the diagonal length of the image. For example, in BSDS500, where images have a resolution of 321$\times$481 pixels, setting $\textit{maxDist} = 0.0075$ results in an error tolerance of approximately 4.3 pixels, computed as $\sqrt{481^2+321^2} \times 0.0075$. Notably, the choice of \textit{maxDist} is often inconsistent across datasets. For instance, in BSDS500\cite{MF2001A}, \textit{maxDist} is set to 0.0075 (about 4.3 pixels), whereas in BIPED\cite{SS2023Dense}, the same value translates to 11.1 pixels due to higher image resolution. In MDBD\cite{MK2016A}, \textit{maxDist} is often increased to 0.011, allowing a tolerance of approximately 16.3 pixels. While a stricter standard (i.e., reducing \textit{maxDist}) would more accurately reflect a model's true localization precision, the necessity of such strict evaluation has not yet been systematically discussed.

\section{Methodology}

This section presents a comprehensive framework for high-precision edge detection (ED). Our approach addresses the limitations of current ED systems through a three-fold synergy: (1) Architectural Design: The proposed SDPED model performs a task-adaptive architectural transfer and modification by re-engineering texture-handling structures from super-resolution (SR) to enhance spatial precision. (2) Learning Paradigm: A novel data augmentation strategy integrates noiseless ideal priors into the training process, breaking the performance ceiling imposed by subjective human annotations. (3) Evaluation Rigor: We analyze the limitations of resolution-dependent error tolerance and discuss fixed-pixel benchmarks as a more rigorous alternative for assessing localization precision.

\subsection{The SDPED Model: Task-adaptive Architectural Transfer and Modification}

The core challenge of high-precision ED lies in the delicate balance between suppressing complex textural regions and accurately localizing true edges. Since textures lack a formal mathematical definition, designing architectures that can distinguish them from edges remains largely an empirical challenge. However, we observe that image super-resolution (SR), a related low-level vision task, has produced numerous architectures experimentally validated for high-frequency feature handling and texture processing. Therefore, transferring and adapting these structural designs to ED tasks represents a reasonable direction. 

\begin{figure}[htbp]
    \centering
    \includegraphics[width=5.5in]{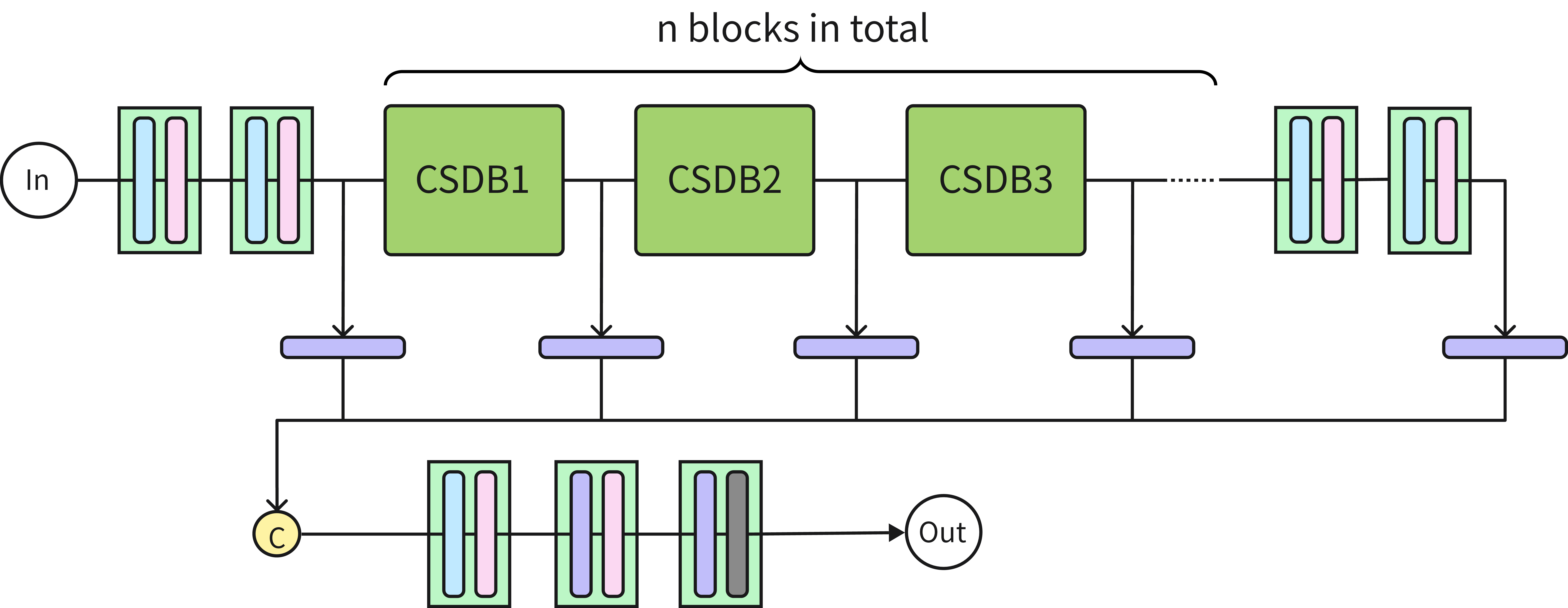}
    \caption{\quad \textbf{Structure of the SDPED Model:} The blue square represents the $3\times 3$ convolutional layer, the purple square represents the $1\times 1$ convolutional layer, the pink square represents the Leaky Relu, the black square represents the Sigmoid, $\oplus$ represents pixel-wise addition, and represents the concatenation. The \textbf{top path} represents the main processing pipeline, where the first two convolution layers output 32 and 64 feature maps, respectively. The last two convolution layers maintain 64 input and output features. The model comprises $n+2$ layers, including the feature extractor block, each CSDB unit, and the final processing block, all of which produce intermediate outputs fused into 21 features. These intermediate outputs are concatenated, resulting in a total of $21 \times (n+2)$ features for the final fusing layer. The \textbf{bottom path} represents the fusing block, where the concatenated features pass through a $3\times 3$ convolution layer followed by two $1 \times 1$ convolution layers. The output feature dimensions of these layers are 256, 512, and 1, respectively.}
    \label{fig:Model}
\end{figure}

\begin{figure}[htbp]
    \centering
    \includegraphics[width=5.5in]{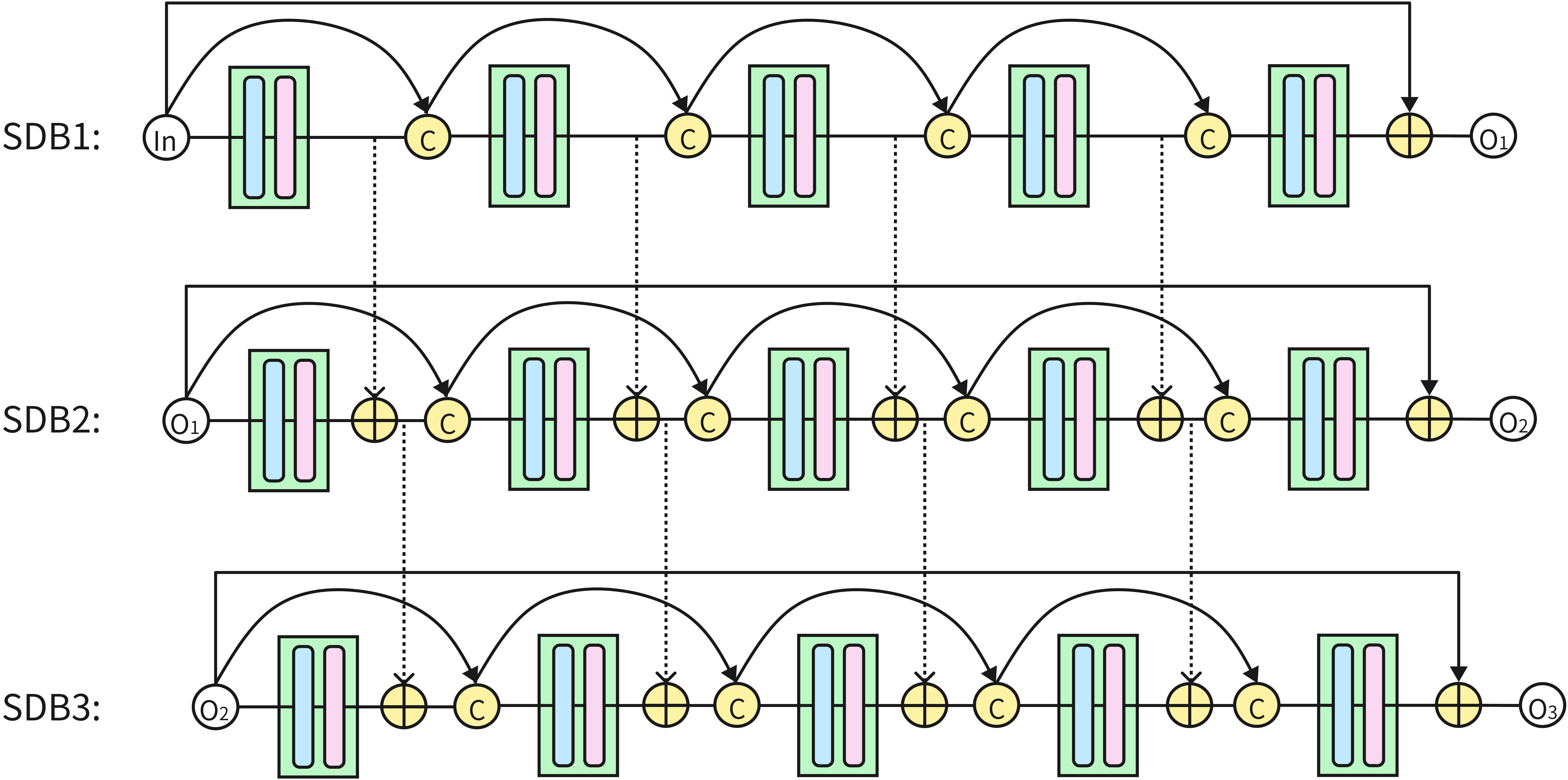}
    \caption{\quad \textbf{Cascading Skipping Dense Block (CSDB):} Each CSDB block consists of three cascaded SDB blocks connected via skip connections. Within a single SDB block, there are five convolution layers. The first four layers output 32 feature maps, while the final layer produces 64 feature maps. The input feature dimensions grow progressively from 64 to 96, 128, 160, and 192 across the first four layers. \textit{In} represents the input, and, $O_{1}$, $O_{2}$ and $O_{3}$ represent the outputs of each SDB, respectively.}
    \label{fig:CSDB}
\end{figure}

Guided by this observation, we propose the SDPED model, as illustrated in Figure \ref{fig:Model}, with structural details provided in Figure \ref{fig:CSDB}. Our model is built upon the Cascaded Skipping Density Block (CSDB), extending the structure of Dense Block (DB) from SR tasks \cite{WY2019ESRGAN, WX2021Real}. While the standard DB has demonstrated exceptional performance in texture handling for SR tasks, our experiments indicate that its direct application to ED leads to suboptimal localization performance; therefore, we reformulate it into the CSDB structure with task-adaptive modifications tailored to ED. Furthermore, we address a common imbalance in existing ED models: many architectures allocate millions of parameters to feature extraction but rely on a simplistic $1\times1$ convolution for feature fusion, which often leads to the underutilization of rich semantic features. To address this, SDPED employs a dedicated, extending fusing block consisting of three convolutional layers (including $3 \times 3$) to enhance feature integration and refine the final edge map, allowing the rich information captured by the CSDBs to be more effectively leveraged for the final prediction.

As in most ED models, the model is trained using the weighted binary cross-entropy (WBCE) loss function, defined as:

\begin{equation}
    L_{WBCE}(\hat{Y},Y)=-\alpha\sum_{y_{i}\in Y^{+}}\log(\hat{y_{i}})-\lambda(1-\alpha)\sum_{y_{i}\in Y^{-}}\log(1-\hat{y_{i}}),
\end{equation}

\noindent where $\hat{Y}$ represents the model's prediction, $Y^{+}$ is the set of positive samples (edge pixels in the ground-truth), $Y^{-}$ is the set of negative samples (non-edge pixels in the ground-truth), and, $y_{i}$ and $\hat{y}_{i}$ represent the corresponding pixels in the ground-truth and prediction, respectively. The weighting factor $\alpha$ is computed as:

\begin{equation}
    \alpha = \frac{|Y^{-}|}{|Y|}.
\end{equation}

\subsection{Ideal Prior Guidance: Enhancing Performance via Noiseless Data} 
\label{sec:Noiseless}

A critical limitation of existing ED models is their inability to yield perfect results even when the input is noiseless (for example, the ground-truth edge map itself). As demonstrated in Fig. \ref{fig:Noiseless}, models produce blurred predictions on noiseless data, similar to their performance on noisy (textured) images. We argue that this stems from the human-annotation ceiling: since models are trained exclusively on noisy, human labels, they are unlikely to learn the ideal physical representation of a noise-free edge.

The most direct solution—obtaining perfectly clean, objective labels—is practically impossible due to the nature of manual annotation. However, we propose a novel paradigm: label-to-data augmentation, which allows noiseless training labels to be added to training sets. Observing that in the context of ED, a ground-truth label can be treated as a noiseless input image because the edge map of an edge map is identical to itself, we provide the model with ideal prior guidance by incorporating ground-truth labels directly into the training set as input samples. Namely, we add noiseless data to the training set by viewing labels as training data too, whose annotations are themselves.

This provides a practically feasible way, and we have not found others yet, to introduce deterministic, noise-free knowledge into the training process without requiring additional human effort. It enables models to learn from perfect data-label pairs on certain samples, thereby enhancing robustness when facing both noisy and noiseless images. Furthermore, this approach is highly generalizable: it can potentially be extended to a wide range of supervised tasks, particularly those where ground-truth labels are inevitably imperfect. As shown in Figure \ref{fig:Noiseless}, this ideal prior training allows models to generate visually superior predictions.

\subsection{Rethinking Evaluation Benchmarks for High-Precision ED}
\label{sec:Benchmark}

To assess model performance, the community widely adopts the methodology proposed in \cite{MF2004Learning}, which computes ODS, OIS, and AP within an error-tolerance distance, denoted by \textit{maxDist}. While the metrics themselves are standardized, the selection criteria for \textit{maxDist} in current research are often inconsistent and lack a rigorous theoretical basis, which poses a significant challenge for high-precision ED tasks.

Currently, \textit{maxDist} is typically defined as a fixed ratio of the image’s diagonal length, which is inherently problematic because it couples the evaluation standard with image resolution. As resolution increases, the error tolerance is relaxed. For instance, a ratio of 0.0075 translates to a 4.3-pixel tolerance for images of resolution $321 \times 481$. However, if one were to concatenate four such images into a single $642 \times 962$ frame, the same ratio rule would double the allowable error to 8.6 pixels, despite the underlying structures remaining identical. Consequently, simply enlarging images can lead to inflated evaluation scores without any actual improvement in detection performance. Such loose constraints mask the deficiencies of models in fine localization, reducing the scientific persuasiveness of the evaluation.

Moreover, the inconsistency across different datasets further complicates the comparability of ED models. Even when using the same ratio, such as 0.0075, the tolerance varies from 4.3 pixels in BRIND ($321 \times 481$) to 11.1 pixels in BIPED ($720 \times 1280$). In the MDBD dataset, the standard is often further relaxed by using a ratio of 0.011, reaching up to 16.3 pixels of error tolerance. This means a prediction that would be penalized as a false positive in one dataset might be accepted as a true positive in another solely due to the resolution-based scaling or relaxed error tolerance, which reduces the persuasiveness of model assessment across datasets.

\begin{figure}[htbp]
    \centering
    \includegraphics[width=4in]{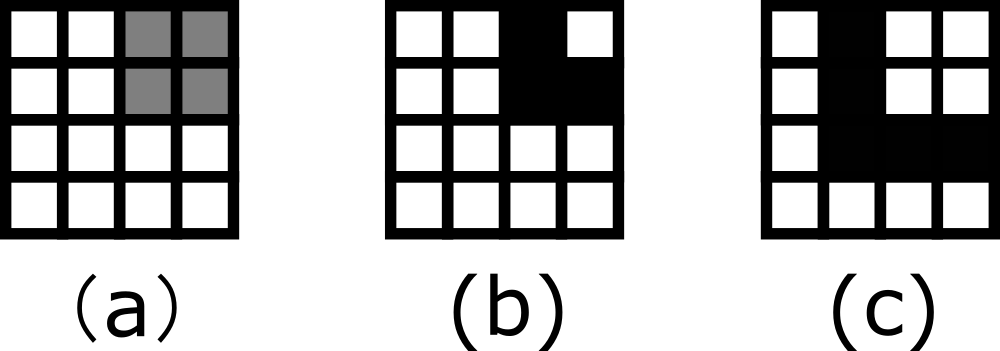}
    \caption{\quad \textbf{Edge annotation inconsistencies:} (a) A typical image region; (b) and (c) illustrate different annotation styles, where black squares denote edge pixels. In (b) and (c), non-corner edge pixels may deviate by one pixel horizontally or vertically, while corner edge pixels may shift diagonally by one pixel.}
    \label{fig:Edges}
\end{figure}

Therefore, a rethinking of the philosophy on using the evaluated algorithm should be conducted. To ensure a resolution-independent evaluation, we propose setting error tolerance as a fixed pixel value rather than a proportional ratio. Furthermore, we contend that models must be evaluated under a low error-tolerance distance, since a high score under a relaxed threshold but a poor one under a strict one indicates that errors are masked and misaligned predictions are contributing disproportionately to the final score\footnote{For example, a prediction missing a single edge pixel should ideally score higher than one missing that pixel while introducing an additional false edge elsewhere. However, a loose error tolerance increases the likelihood of the latter receiving a higher score, as the erroneous pixel may still fall within the large tolerance range.}. However, the strictest possible standard—a zero-pixel tolerance—is often unrealistic due to the inherent nature of human labels. As illustrated in Figure \ref{fig:Edges}, edge annotations in current datasets are manually labeled and often exhibit a one-pixel deviation. In physical images, a true edge typically occupies a two-pixel width (spanning the boundary between two regions), yet ground-truth labels are usually marked as a one-pixel-wide path. This subjective choice means labels may vary by up to one pixel horizontally, vertically, or diagonally, depending on the annotator's preference. Therefore, our analysis suggests that zero-pixel tolerance is overly punitive toward legitimate annotation inconsistencies, while relaxed error tolerance risks accepting erroneous noise. Based on these analyses, we identify the 1-pixel threshold as the strictest standard for evaluation, in practice.

\section{Experiment}
\label{sec:Experiment}

This section presents the experimental validation of our framework. We first detail the dataset configurations and training protocols, followed by a description of the evaluation benchmarks. Finally, we provide experimental results and ablation studies to validate the effectiveness of the SDPED architecture and justify the proposed ideal prior guidance training as well as the stricter evaluation benchmark.

\subsection{Data and Training Setting}

This subsection presents the dataset settings and training methods.

\subsubsection{Dataset}

Our experiments were conducted on four datasets: BRIND\cite{PH2021RINDNet}, UDED\cite{SL2023Tiny}, MDBD\cite{MK2016A}, and BIPED\cite{SR2020Dense}. BRIND provides multiple annotations per image, from which we synthesize a unified binary ground-truth by marking a pixel as an edge if it is identified by any annotator. For MDBD, we use annotation 3 as the ground-truth\footnote{Annotation 3 is chosen arbitrarily; alternative combinations of the six available annotations were tested, but the resulting ground-truths were found to be less reliable.}. UDED and BIPED each provide a single label per image, which is used as the ground-truth. Notably, among these datasets, UDED exhibits the highest annotation quality, while BRIND offers the most diverse image collection.

\subsubsection{Data Augmentation}

To enhance data diversity, training images are recursively split in half until their height and width are both below 640 pixels. Each resulting sub-image undergoes eight augmentation transformations: rotations of 0$^{o}$, 90$^{o}$, 180$^{o}$, and 270$^{o}$, combined with horizontal flips. Furthermore, to implement the ideal prior guidance discussed in Section \ref{sec:Noiseless}, we augment the training set by incorporating ground-truth edge maps as noiseless input samples, where each label serves as both the input and the target.

\subsubsection{Training Detail}

All models are trained using the Adam optimizer with a weight decay of $10^{-8}$. The initial learning rate is set to $10^{-4}$ and is reduced by a factor of 10 every 50 epochs. The batch size is fixed at 8 or 4, and the balance parameter in the loss function is set to $\lambda = 1.1$ (please refer to Equation (4)). During training, images are randomly cropped to 320$\times$320 pixels, and the dataset is refreshed every five epochs. To avoid resizing operations during training, the two UDED images with widths below 320 pixels are designated as test images. The number of training epochs is set as follows: BRIND: 100 epochs, UDED: 150 epochs, MDBD: 100 epochs, BIPED: 75 epochs.

\subsection{Benchmark}

Evaluation benchmarks are clarified in this subsection.

\subsubsection{Evaluation Settings}

All models are evaluated on full-resolution test images\footnote{Unless explicitly stated, noiseless data are excluded from testing.}. The standard evaluation algorithm from \cite{MF2004Learning} is used, under the strictest 1-pixel criterion (1 pixel in both horizontal, vertical, and diagonal directions) proposed in Section \ref{sec:Benchmark}. The \textit{maxDist} parameter is calibrated for each dataset (BRIND: 0.003, MDBD/BIPED: 0.001, UDED: 0.004 (1-pixel tolerance for the smallest image)) to ensure a consistent 1-pixel error tolerance in the diagonal direction across varying resolutions.

As noted in \cite{MK2022Have}, many models achieve SOTA performance on a specific dataset at the cost of generalization, leading to significant performance degradation when tested on different datasets\footnote{This issue can arise even when only the training-test partitioning of the same dataset is altered.}. To ensure robustness and reduce partition-specific bias, we conduct experiments on multiple datasets with various dataset partitions in testing the SDPED architecture. We use the notation \textit{'Dataset-$P_{k}$-a-b-Ec'} to index dataset partitions, where: \textit{Dataset} refers to the dataset name, \textit{$P_{k}$} denotes the $k$-th partition, \textit{$a$} and \textit{$b$} represent the number of training and test images, respectively, and \textit{$c$} indicates the number of training epochs. For example, \textit{'BRIND-$P_{1}$-250-250-E100'} refers to experiments conducted on the BRIND dataset using the first partition, where 250 images are used for training, 250 for testing, and the model is trained for 100 epochs.

\subsubsection{Comparison Settings}

The SDPED model is compared against several methods, including CNN-based and transformer-based: HED\cite{XT2015Holistically}, RCF\cite{LC2017Richer}, RIND \cite{PH2021RINDNet}, BDCN\cite{HZ2022BDCN}, Dexi\cite{SS2023Dense}, EdgeNat\cite{JG2024EdgeNAT}, and daf\cite{WL2025Enhanced}. Since the SDPED architecture can be configured with different numbers of CSDB blocks, we denote a model with $n$ CSDB blocks as SDPED$\_n$, with SDPED$\_7$ being the standard for comparison. To ensure a rigorous and fair comparison, all models are retrained from scratch under identical conditions, utilizing the same unified training-test partitions, optimizer settings, and evaluation pipeline.

\subsection{Experiment Results}

Experiment results are provided in this subsection. First, experiments comparing SDPED to previous models and ablation studies on the SDPED architecture are conducted, using the proposed data augmentation method and the evaluation criterion. Then, justification experiments for those approaches are presented.

\subsubsection{Experiments and ablation studies for SDPED}

\begin{figure}[htbp]
	\centering
	\includegraphics[width=5.5in]{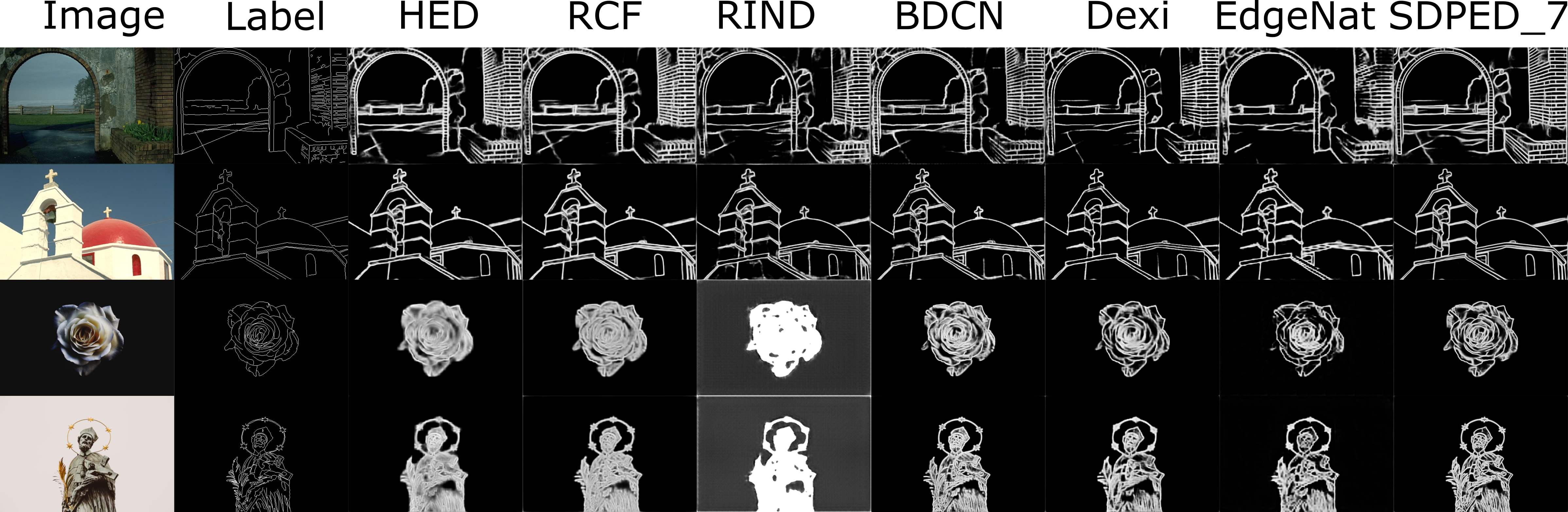}
	\caption{\quad Sample predictions on BRIND and UDED, all from partition $P_{1}$ and displayed without NMS. The results illustrate that our model generates more visually coherent and spatially consistent predictions.}
	\label{fig:Compare}
\end{figure}

The quantitative results validating the effectiveness of SDPED on the BRIND, UDED, MDBD, and BIPED2 datasets are summarized in Tables \ref{BRIND} to \ref{BIPED2}, with qualitative visual comparisons provided in Figure \ref{fig:Compare}. Following the evaluation protocol, the best and second-best performances are highlighted in \textcolor{blue}{blue} and \textcolor{red}{red}, respectively. The bottom row of each table quantifies the performance margin achieved by SDPED$_{7}$ relative to the previous SOTA.

Our framework obtains consistent improvements compared to previous methods, with particularly notable gains in AP. On BRIND, SDPED$_{7}$ achieves average gains of approximately 2.7\% in ODS, 3.2\% in OIS, and 6.9\% in AP. On UDED, the improvements are approximately 5.2\% in ODS, 4.4\% in OIS, and 2.4\% in AP\footnote{It is worth noting that certain baseline models may exhibit significant performance fluctuations across different dataset partitions; while RIND performing reasonably on $P_{3}$, it yielded significantly lower scores on $P_{1}$ and $P_{2}$. When retraining RIND three times on $P_{2}$, it consistently yielded low scores, while retraining it on $P_{1}$ three times resulted in twice unsatisfactory outcomes. This empirical observation indicates that certain architectures may be sensitive to data partitioning, further motivating the use of multi-partition evaluations. This issue also appears on the dataset MDBD.}. On MDBD, under both the strictest (0.001 or 1-pixel) and relaxed (0.003 or 4.4 pixels) error tolerances, our model demonstrates superior improvements in AP of about 22.5\% and 14.7\%, respectively, while also improving ODS/OIS by about $4.8\%/4.4\%$ and $4.8\%/2.9\%$ on the two criteria, respectively. On BIPED2, while SDPED$_{7}$ achieves comparable ODS and OIS results (within 0.3\% margin on average), it delivers a significant 9.52\% improvement in AP.

\begin{table}[htbp]\scriptsize
\renewcommand\arraystretch{1.5} 
\centering
\caption{\qquad \textbf{Results on BRIND with error toleration 0.003 (1 pixel on the diagonal) with NMS:} Across all the experiments and benchmarks, our model (SDPED$\_7$) achieves the best results. On average, our model improves scores by approximately $2.7\%$ in ODS, $3.2\%$ in OIS, and $6.9\%$ in AP.}
\label{BRIND}
\centering

\begin{tabular}{|p{20mm}<{\centering}|p{32.75mm}<{\centering}|p{32.75mm}<{\centering}|p{32.75mm}<{\centering}|}
\hline
Partitions & BRIND-$P_{1}$-250-250-E100 & BRIND-$P_{2}$-300-200-E100 & BRIND-$P_{3}$-400-100-E100 \\
\end{tabular}
\begin{tabular}{|p{20mm}<{\centering}|p{8mm}<{\centering}|p{8mm}<{\centering}|p{8mm}<{\centering}|p{8mm}<{\centering}|p{8mm}<{\centering}|p{8mm}<{\centering}|p{8mm}<{\centering}|p{8mm}<{\centering}|p{8mm}<{\centering}|}
\hline
Benchmarks   & ODS   & OIS   & AP  & ODS   & OIS   & AP  & ODS   & OIS   & AP \\
\hline
HED & 0.639 & 0.651 & \textcolor{red}{0.575} & 0.657 & 0.669 & 0.597 & 0.663 & \textcolor{red}{0.675} & \textcolor{red}{0.605} \\
\hline
RCF & \textcolor{red}{0.643} & \textcolor{red}{0.653} & 0.573 & \textcolor{red}{0.663} & \textcolor{red}{0.672} & 0.598 & \textcolor{red}{0.665} & 0.674 & 0.603 \\
\hline
RIND & 0.624 & 0.634 & 0.509 & 0.651 & 0.661 & \textcolor{red}{0.610} & 0.649 & 0.662 & 0.590 \\
\hline
BDCN & 0.639 & 0.652 & 0.567 & 0.645 & 0.659 & 0.576 & 0.662 & 0.673 & 0.587 \\
\hline
Dexi & 0.639 & 0.648 & 0.548 & 0.652 & 0.663 & 0.572 & 0.663 & 0.674 & 0.585 \\
\hline
EdgeNat & 0.577 & 0.586 & 0.452 & 0.585 & 0.594 & 0.457& 0.597 & 0.606 & 0.473 \\
\hline
daf & 0.635 & 0.644 & 0.567 & 0.646 & 0.661 & 0.585& 0.655 & 0.665 & 0.582 \\
\hline
SDPED$\_7$(Ours) & \textcolor{blue}{\textbf{0.660}} & \textcolor{blue}{\textbf{0.672}} & \textcolor{blue}{\textbf{0.613}} & \textcolor{blue}{\textbf{0.677}} & \textcolor{blue}{\textbf{0.693}} & \textcolor{blue}{\textbf{0.643}} & \textcolor{blue}{\textbf{0.687}} & \textcolor{blue}{\textbf{0.698}} & \textcolor{blue}{\textbf{0.658}} \\
\hline
Improve & 2.64\% & 2.91\% & 6.61\% & 2.11\% & 3.13\% & 5.41\%& 3.31\% & 3.41\% & 8.76\% \\
\hline
\end{tabular}
\begin{tabular}{|p{131.25mm}<{\centering}|}
The results of the SDPED model with lengths 5 and 3 are listed below \\
\end{tabular}
\begin{tabular}{|p{20mm}<{\centering}|p{8mm}<{\centering}|p{8mm}<{\centering}|p{8mm}<{\centering}|p{8mm}<{\centering}|p{8mm}<{\centering}|p{8mm}<{\centering}|p{8mm}<{\centering}|p{8mm}<{\centering}|p{8mm}<{\centering}|}
\hline
SDPED$\_5$(Ours) & 0.660 & 0.672 & 0.616 & 0.676 & 0.692 & 0.645 & 0.683 & 0.693 & 0.649 \\
\hline
SDPED$\_3$(Ours) & 0.654 & 0.668 & 0.617 & 0.674 & 0.687 & 0.647 & 0.684 & 0.695 & 0.658 \\
\hline
\end{tabular}
\end{table}

\begin{table}[htbp]\scriptsize
\renewcommand\arraystretch{1.5}
\centering
\caption{\qquad \textbf{Results on UDED with an error tolerance of 0.004 (1 pixel along the diagonal for the lowest-resolution image) with NMS:} Across all three experiments and benchmarks, our model consistently achieves the best performance. On average, our model improves scores by approximately $5.2\%$ in ODS, $4.4\%$ in OIS, and $2.4\%$ in AP.}
\label{UDED}

\begin{tabular}{|p{20mm}<{\centering}|p{32.75mm}<{\centering}|p{32.75mm}<{\centering}|p{32.75mm}<{\centering}|}
\hline
\ Partitions &  UDED-$P_{1}$-18-11-E150 &  UDED-$P_{2}$-20-9-E150 & UDED-$P_{3}$-13-14-E150
\end{tabular}
\begin{tabular}{|p{20mm}<{\centering}|p{8mm}<{\centering}|p{8mm}<{\centering}|p{8mm}<{\centering}|p{8mm}<{\centering}|p{8mm}<{\centering}|p{8mm}<{\centering}|p{8mm}<{\centering}|p{8mm}<{\centering}|p{8mm}<{\centering}|}
\hline
\ Benchmarks   & ODS   & OIS   & AP  & ODS   & OIS   & AP  & ODS   & OIS   & AP \\
\hline
HED & 0.714 & 0.732 & 0.687 & 0.709 & 0.733 & 0.673 & 0.639 & 0.675 & 0.590 \\
\hline
RCF & 0.737 & 0.775 & 0.713 & 0.746 & 0.763 & 0.724 & 0.637 & 0.703 & 0.609 \\
\hline
RIND & 0.416 & 0.425 & 0.071 & 0.375 & 0.381 & 0.046 & \textcolor{red}{0.716} & 0.747 & 0.678 \\
\hline
BDCN & 0.767 & \textcolor{red}{0.787} & 0.763 & \textcolor{red}{0.772} & \textcolor{red}{0.803} & 0.757 & 0.689 & \textcolor{red}{0.762} & 0.686 \\
\hline
Dexi & \textcolor{red}{0.771} & 0.784 & \textcolor{red}{0.766} & 0.767 & 0.793 & \textcolor{red}{0.779} & 0.701 & 0.750 & \textcolor{red}{0.731} \\
\hline
EdgeNat  & 0.709 & 0.714 & 0.601 & 0.710 & 0.724 & 0.633& 0.678 & 0.695 & 0.591 \\
\hline
daf & 0.750 & 0.776 & 0.719 & 0.757 & 0.786 & 0.727& 0.688 & 0.748 & 0.650 \\
\hline
SDPED$\_7$(Ours) & \textcolor{blue}{\textbf{0.812}} & \textcolor{blue}{\textbf{0.828}} & \textcolor{blue}{\textbf{0.792}} & \textcolor{blue}{\textbf{0.807}} & \textcolor{blue}{\textbf{0.830}} & \textcolor{blue}{\textbf{0.791}} & \textcolor{blue}{\textbf{0.757}} & \textcolor{blue}{\textbf{0.798}} & \textcolor{blue}{\textbf{0.748}} \\
\hline
Improve & 5.32\% & 5.21\% & 3.39\% & 4.53\% & 3.36\% & 1.54\%& 5.73\% & 4.72\% & 2.33\% \\
\hline
\end{tabular}
\begin{tabular}{|p{131.25mm}<{\centering}|}
The results of the SDPED model with lengths 5 and 3 are listed below \\
\end{tabular}
\begin{tabular}{|p{20mm}<{\centering}|p{8mm}<{\centering}|p{8mm}<{\centering}|p{8mm}<{\centering}|p{8mm}<{\centering}|p{8mm}<{\centering}|p{8mm}<{\centering}|p{8mm}<{\centering}|p{8mm}<{\centering}|p{8mm}<{\centering}|}
\hline
SDPED$\_5$(Ours) & 0.811 & 0.824 & 0.795 & 0.806 & 0.827 & 0.796 & 0.756 & 0.793 & 0.744 \\
\hline
SDPED$\_3$(Ours) & 0.807 & 0.822 & 0.786 & 0.803 & 0.825 & 0.798 & 0.753 & 0.791 & 0.751 \\
\hline
\end{tabular}
\end{table}

\begin{table}[htbp]\scriptsize
	\renewcommand\arraystretch{1.5}
	\centering
	\caption{\qquad \textbf{Results on MDBD with error tolerances of 0.001 (1.5 pixels in the horizontal and vertical directions, respectively) with NMS:} On average, our model improves scores by approximately 4.8$\%$ in ODS, 4.8$\%$ in OIS, and 22.5$\%$ in AP, respectively.}
	\label{MDBD1}
	\begin{tabular}{|p{17mm}<{\centering}|p{32.75mm}<{\centering}|p{32.75mm}<{\centering}|p{32.75mm}<{\centering}|}
		\hline
		\ Partitions & MDBD-$P_{1}$-60-40-E100  & MDBD-$P_{2}$-70-30-E100  & MDBD-$P_{3}$-80-20-E100
	\end{tabular}
	\begin{tabular}{|p{17mm}<{\centering}|p{8mm}<{\centering}|p{8mm}<{\centering}|p{8mm}<{\centering}|p{8mm}<{\centering}|p{8mm}<{\centering}|p{8mm}<{\centering}|p{8mm}<{\centering}|p{8mm}<{\centering}|p{8mm}<{\centering}|}
		\hline
		\ Benchmarks   & ODS   & OIS   & AP  & ODS   & OIS   & AP  & ODS   & OIS   & AP \\
		\hline
		HED  & 0.302 & 0.304 & 0.122 & 0.289 & 0.291 & 0.114 & 0.261 & 0.264 & 0.094\\
		\hline
		RCF & 0.276 & 0.277 & 0.096 & 0.264 & 0.266 & 0.089 & 0.245 & 0.246 & 0.079\\
		\hline
		RIND & 0.243 & 0.244 & 0.035 & 0.287 & 0.290 & \textcolor{red}{0.118} & \textcolor{red}{0.285} & \textcolor{red}{0.292} & \textcolor{red}{0.125} \\
		\hline
		BDCN & 0.277 & 0.278 & 0.104 & 0.270 & 0.273 & 0.097 & 0.255 & 0.257 & 0.088\\
		\hline
		Dexi & \textcolor{red}{0.305} & \textcolor{red}{0.307} & \textcolor{red}{0.126} & \textcolor{red}{0.294} & \textcolor{red}{0.297} & 0.116 & 0.271 & 0.274 & 0.102 \\
		\hline
        EdgeNat & 0.277 & 0.279 & 0.101 & 0.247 & 0.249 & 0.081& 0.243 & 0.245 & 0.082 \\
		\hline
        daf & 0.293 & 0.296 & 0.116 & 0.266 & 0.268 & 0.091& 0.245 & 0.250 & 0.084 \\
		\hline
		SDPED$\_7$ & \textcolor{blue}{\textbf{0.327}} & \textcolor{blue}{\textbf{0.330}} & \textcolor{blue}{\textbf{0.165}} & \textcolor{blue}{\textbf{0.309}} & \textcolor{blue}{\textbf{0.312}} & \textcolor{blue}{\textbf{0.148}} & \textcolor{blue}{\textbf{0.291}} & \textcolor{blue}{\textbf{0.297}} & \textcolor{blue}{\textbf{0.139}} \\
		\hline
		Improve & 7.21\% & 7.49\% & 30.95\% & 5.10\% & 5.05\% & 25.42\%& 2.11\% & 1.71\% & 11.20\% \\
		\hline
	\end{tabular}
\end{table}

\begin{table}[htbp]\scriptsize
	\renewcommand\arraystretch{1.5}
	\centering
	\caption{\qquad \textbf{Results on MDBD with error 0.003 (4.4 pixels in the horizontal and vertical directions, respectively) with NMS:} Since all models exhibit limited absolute performance under the strictest benchmark (0.001 error tolerance), which may be influenced by annotation quality, to provide a broader evaluation, we also assess the performance using a more relaxed standard (0.003 error tolerance or 4.4 pixels). On average, our model improves scores by approximately 2.4$\%$ in ODS, 2.9$\%$ in OIS, and 14.7$\%$ in AP, respectively.}
	\label{MDBD2}
	\begin{tabular}{|p{17mm}<{\centering}|p{32.75mm}<{\centering}|p{32.75mm}<{\centering}|p{32.75mm}<{\centering}|}
		\hline
		\ Partitions & MDBD-$P_{1}$-60-40-E100  & MDBD-$P_{2}$-70-30-E100  & MDBD-$P_{3}$-80-20-E100
	\end{tabular}
	\begin{tabular}{|p{17mm}<{\centering}|p{8mm}<{\centering}|p{8mm}<{\centering}|p{8mm}<{\centering}|p{8mm}<{\centering}|p{8mm}<{\centering}|p{8mm}<{\centering}|p{8mm}<{\centering}|p{8mm}<{\centering}|p{8mm}<{\centering}|}
		\hline
		\ Benchmarks   & ODS   & OIS   & AP  & ODS   & OIS   & AP  & ODS   & OIS   & AP \\
		\hline
		HED  & 0.565 & 0.570 & 0.425 & 0.550 & 0.554 & 0.396 & 0.515 & 0.520 & 0.356\\
		\hline
		RCF & 0.538 & 0.540 & 0.356 & 0.525 & 0.527 & 0.339 & 0.492 & 0.495 & 0.307\\
		\hline
		RIND & 0.455 & 0.460 & 0.091 & 0.555 & \textcolor{red}{0.564} & \textcolor{red}{0.438} & \textcolor{red}{0.531} & \textcolor{red}{0.550} & \textcolor{red}{0.445} \\
		\hline
		BDCN & 0.542 & 0.545 & 0.377 & 0.533 & 0.536 & 0.362 & 0.514 & 0.515 & 0.339\\
		\hline
		Dexi & \textcolor{red}{0.571} & \textcolor{red}{0.576} & \textcolor{red}{0.433} & \textcolor{red}{0.558} & 0.563 & 0.407 & 0.521 & 0.529 & 0.374 \\
		\hline
        EdgeNat & 0.544 & 0.547 & 0.370 & 0.498 & 0.501 & 0.317& 0.490 & 0.493 & 0.314 \\
		\hline
        daf & 0.553 & 0.558 & 0.406 & 0.522 & 0.524 & 0.335& 0.489 & 0.495 & 0.316 \\
		\hline
		SDPED$\_7$ & \textcolor{blue}{\textbf{0.590}} & \textcolor{blue}{\textbf{0.599}} & \textcolor{blue}{\textbf{0.531}} & \textcolor{blue}{\textbf{0.569}} & \textcolor{blue}{\textbf{0.579}} & \textcolor{blue}{\textbf{0.501}} & \textcolor{blue}{\textbf{0.543}} & \textcolor{blue}{\textbf{0.561}} & \textcolor{blue}{\textbf{0.477}} \\
		\hline
		Improve & 3.33\% & 3.99\% & 22.63\% & 1.97\% & 2.66\% & 14.38\%& 2.26\% & 2.00\% & 7.19\% \\
		\hline
	\end{tabular}
\end{table}

\begin{table}[htbp]\scriptsize
	\renewcommand\arraystretch{1.3}
	\centering
	\caption{\qquad \textbf{Results on BIPED2 with error tolerance 0.001 (1 pixel along the diagonal) with NMS:} The last part presents the average results across the three partitions. Compared to previous approaches, SDPED$_7$  obtains significant superiority in AP, improving by approximately 9.5$\%$, and comparable performance on ODS and OIS, differing less than 0.3$\%$, on average.}
	\label{BIPED2}
	\begin{tabular}{|p{25mm}<{\centering}|p{41.75mm}<{\centering}|p{41.75mm}<{\centering}|}
		\hline
		\ Partitions &  BIPED2-$P_{1}$-125-125-E75 &  BIPED2-$P_{2}$-175-75-E75
	\end{tabular}
	\begin{tabular}{|p{25mm}<{\centering}|p{11mm}<{\centering}|p{11mm}<{\centering}|p{11mm}<{\centering}|p{11mm}<{\centering}|p{11mm}<{\centering}|p{11mm}<{\centering}|}
		\hline
		\ Benchmarks   & ODS   & OIS   & AP  & ODS   & OIS   & AP\\
		\hline
		HED & 0.640 & 0.643 & 0.493 & 0.645 & 0.648 & 0.496\\
		\hline
		RCF & 0.641 & 0.644 & 0.509 & 0.644 & 0.647 & 0.510\\
		\hline
		RIND & 0.673 & 0.674 & 0.479 & \textcolor{blue}{\textbf{0.688}} & \textcolor{red}{\textbf{0.691}} & 0.468\\
		\hline
		BDCN & 0.654 & 0.657 & 0.519 & 0.667 & 0.670 & 0.530\\
		\hline
		Dexi & \textcolor{red}{0.674} & 0.675 & 0.530 & 0.683 & 0.685 & 0.544\\
		\hline
        EdgeNat & 0.590 & 0.592 & 0.423 & 0.601 & 0.603 & 0.438\\
		\hline
		 daf & \textcolor{blue}{0.675} & \textcolor{red}{0.676} & \textcolor{red}{0.533} & 0.692 & \textcolor{blue}{0.693} & \textcolor{red}{0.569}\\
		\hline
        SDPED$\_7$ & \textcolor{blue}{\textbf{0.675}} & \textcolor{blue}{\textbf{0.678}} & \textcolor{blue}{\textbf{0.599}} & \textcolor{red}{0.684} & 0.687 & \textcolor{blue}{\textbf{0.608}}\\
		\hline
		Improve & 0.00\% & 0.30\% & 12.38\% & -0.58\% & 0.29\% & 6.85\%\\
        \hline
	\end{tabular}
    \begin{tabular}{|p{25mm}<{\centering}|p{41.75mm}<{\centering}|p{41.75mm}<{\centering}|}
		\hline
		 & BIPED2-$P_{3}$-200-50-E75 & Average
	\end{tabular}
	\begin{tabular}{|p{25mm}<{\centering}|p{11mm}<{\centering}|p{11mm}<{\centering}|p{11mm}<{\centering}|p{11mm}<{\centering}|p{11mm}<{\centering}|p{11mm}<{\centering}|}
		\hline
		HED & 0.640 & 0.644 & 0.494 & 0.642 & 0.645 & 0.494\\
		\hline
		RCF & 0.640 & 0.644 & 0.497 & 0.642 & 0.645 & 0.505\\
		\hline
		RIND & \textcolor{blue}{\textbf{0.677}} & \textcolor{blue}{\textbf{0.681}} & 0.341 & \textcolor{blue}{\textbf{0.679}} & \textcolor{blue}{\textbf{0.682}} & 0.429\\
		\hline
		BDCN & 0.655 & 0.659 & 0.512 & 0.659 & 0.662 & 0.520 \\
		\hline
		Dexi & \textcolor{blue}{\textbf{0.677}} & 0.679 & 0.530 & \textcolor{red}{0.678} & \textcolor{red}{0.680} & 0.535 \\
		\hline
        EdgeNat & 0.575 & 0.579 & 0.408 & 0.589 & 0.591 & 0.423 \\
		\hline
		daf & 0.663 & 0.666 & \textcolor{red}{0.535} & 0.677 & 0.678 & \textcolor{red}{0.546}\\
		\hline
        SDPED$\_7$ & \textcolor{red}{0.673} & \textcolor{red}{0.680} & \textcolor{blue}{\textbf{0.588}} & 0.677 & \textcolor{blue}{\textbf{0.682}} & \textcolor{blue}{\textbf{0.598}} \\
		\hline
		Improve & -0.59\% & -0.15\% & 9.91\% & -0.15\% & 0.29\% & 9.52\%\\
		\hline
	\end{tabular}
\end{table}

We conducted ablation studies on the BRIND, UDED, and BIPED2 datasets to verify the contributions of our key architectural components: the CSDB structure and the multi-layer fusing block. As shown in Table \ref{Ablation}, both components contribute positively and consistently to performance across datasets.

\begin{table}[htbp]\scriptsize
	\renewcommand\arraystretch{1.5} 
	\centering
	\caption{\textbf{Ablation study of SDPED$\_7$ on BRIND, UDED, and BIPED2:} \textit{No skipping} refers to removing all skipping connections in SDB (excluding residual ones), while \textit{Single fuse} replaces the multiple-layer fusion block with a single $1 \times 1$ convolution layer. \textit{No skipping with single fuse} removes all skipping connections and applies a single $1 \times 1$ convolution layer for fusion, both as above. The best results are highlighted in \textbf{bold}.}
	\label{Ablation}
	\begin{tabular}{|p{38mm}<{\centering}|p{26.725mm}<{\centering}|p{26.725mm}<{\centering}|p{26.725mm}<{\centering}|}
		\hline
		Partitions & BRIND-$P_{1}$ & UDED-$P_{2}$ & BIPED2-$P_{3}$
	\end{tabular}
	\begin{tabular}{|p{38mm}<{\centering}|p{6mm}<{\centering}|p{6mm}<{\centering}|p{6mm}<{\centering}|p{6mm}<{\centering}|p{6mm}<{\centering}|p{6mm}<{\centering}|p{6mm}<{\centering}|p{6mm}<{\centering}|p{6mm}<{\centering}|}
		\hline
		Benchmark & ODS & OIS & AP & ODS & OIS & AP & ODS & OIS & AP \\
		\hline
		SDPED$\_7$ & \textbf{0.660} & \textbf{0.672} & \textbf{0.613} & \textbf{0.807} & \textbf{0.830} & \textbf{0.791} & \textbf{0.677} & \textbf{0.682} & \textbf{0.598} \\
		\hline
		No skipping & 0.654 & 0.664 & 0.589 & 0.776 & 0.803 & 0.772 & 0.670 & 0.676 & 0.577 \\
		\hline
		Single fuse & 0.657 & 0.669 & 0.609 & 0.798 & 0.820 & 0.786 & 0.664 & 0.669 & 0.568 \\
		\hline
		No skipping with single fuse & 0.649 & 0.660 & 0.581 & 0.786 & 0.812 & 0.751 & 0.666 & 0.670 & 0.565 \\
		\hline
	\end{tabular}
\end{table}

\subsubsection{Justification of the ideal prior guidance}

Here, we justify the ideal prior guidance learning by experiments on various data augmentation methods. Six types of data augmentation methods are applied to training BDCN, Dexi, EdgeNat, daf, and SDPED$_{7}$, and both performances on ordinary test images and noiseless test images are evaluated. In the following tables, \textbf{\textit{Full}} represents the data augmentation with crops, 4 angles of rotation, and flips, \textbf{\textit{Middle}} represents the data augmentation with crops and 4 angles of rotation, but without flips, and \textbf{\textit{Small}} represents only applying crops without rotations or flips. Moreover, \textbf{\textit{+N}} represents the application of adding noiseless data to the training set as suggested in the methodology section, while \textbf{\textit{without +N}} represents not applying noiseless data as previous works did. The evaluation protocol follows the one used in the previous section.

Tables \ref{BRIND-aug} to \ref{BIPED2-aug} display the experiment results on partition 1, which contains the least number of training data for BRIND, MDBD, and BIPED2, making it easier to observe the influence of various augmentation methods\footnote{We do not use the partition with the least number of training images for UDED because training a model requires certain number of images but UDED has too few image unless augmentation significantly.}. Results show that, compared to data augmentation without noiseless data, all the data augmentation methods with noiseless data, as suggested in the methodology section, obtain comparable or better performance on ordinary test images while obtaining significantly better results on noiseless test data, both for ODS, OIS, and AP\footnote{For evaluation across noiseless data, normalized AP, defined by the ordinary AP divided by the length of the interval of recall rate, is used instead of ordinary AP (also denoted by \textbf{\textit{AP}}). This is because the ordinary AP calculates the integral of precision rate over the range of recall rate, which is typically assumed to be [0,1], but for evaluation on noiseless data, the recall rate can always be high regardless of the binary threshold choice in the evaluation, which can result in a very low AP even when the precision rate is always high since the length of the integral interval might be very short. Therefore, the AP benchmark, as it is used to measure the average precision across recalls, should be normalized. However, this is not a problem for evaluation across ordinary test images because the interval of the recall rate is [0,1] mostly. Therefore, for evaluation on ordinary images, we use ordinary AP to be consistent with previous works, while for assessment on noiseless data, we use normalized AP instead.}. This supports the justification of including ideal prior guidance in model training.  

\begin{table}[htbp]\scriptsize
\renewcommand\arraystretch{1.25} 
\centering
\caption{\qquad \textbf{Results with error toleration 0.003 on BRIND-$P_{1}$-250-250.} \textbf{\textit{Avg}} represents the average performance among models with the same data augmentation.}
\label{BRIND-aug}
\centering

\begin{tabular}{|p{32mm}<{\centering}|p{8mm}<{\centering}|p{8mm}<{\centering}|p{8mm}<{\centering}|p{8mm}<{\centering}|p{8mm}<{\centering}|p{8mm}<{\centering}|}
\hline
&\multicolumn{3}{|c|}{Test on Ordinary Data}&\multicolumn{3}{|c|}{Test on Noiseless Data}\\
\hline
Benchmarks   & ODS   & OIS   & AP  & ODS   & OIS   & AP \\
\hline
BDCN(Full+N) & 0.639 & 0.652 & 0.567 & 0.892 & 0.896 & 0.938   \\
BDCN(Full) & 0.635 & 0.647 & 0.559 & 0.716 & 0.725 & 0.748   \\
BDCN(Middle+N) & 0.619 & 0.632 & 0.561 & 0.949 & 0.953 & 0.985   \\
BDCN(Middle) & 0.638 & 0.652 & 0.589 & 0.733 & 0.747 & 0.779   \\
BDCN(Small+N)   & 0.587 & 0.606 & 0.550& 0.980 & 0.983 & 0.998 \\
BDCN(Small)   & 0.593 & 0.614 & 0.558& 0.733 & 0.751 & 0.787 \\
\hline
Dexi(Full+N) & 0.639 & 0.648 & 0.548 & 0.908 & 0.913 & 0.965   \\
Dexi(Full) & 0.627 & 0.636 & 0.535 & 0.748 & 0.752 & 0.751   \\
Dexi(Middle+N) & 0.628 & 0.638 & 0.544 & 0.944 & 0.947 & 0.986   \\
Dexi(Middle) & 0.613 & 0.622 & 0.525 & 0.776 & 0.780 & 0.820   \\
Dexi(Small+N)   & 0.548 & 0.555 & 0.463& 0.954 & 0.956 & 0.990 \\
Dexi(Small)   & 0.471 & 0.479 & 0.393& 0.568 & 0.573 & 0.556 \\
\hline
EdgeNat(Full+N) & 0.577 & 0.586 & 0.452 & 0.942 & 0.945 & 0.984   \\
EdgeNat(Full) & 0.565 & 0.575 & 0.424 & 0.812 & 0.814 & 0.923   \\
EdgeNat(Middle+N) & 0.580 & 0.588 & 0.465 & 0.911 & 0.911 & 0.973   \\
EdgeNat(Middle) & 0.573 & 0.582 & 0.457 & 0.870 & 0.872 & 0.972   \\
EdgeNat(Small+N)   & 0.560 & 0.568 & 0.452& 0.906 & 0.907 & 0.961 \\
EdgeNat(Small)   & 0.556 & 0.566 & 0.457& 0.820 & 0.822 & 0.923 \\
\hline
daf(Full+N) & 0.635 & 0.644 & 0.567 & 0.887 & 0.890 & 0.928   \\
daf(Full) & 0.645 & 0.657 & 0.589 & 0.690 & 0.700 & 0.743   \\
daf(Middle+N) & 0.622 & 0.635 & 0.575 & 0.922 & 0.926 & 0.951   \\
daf(Middle) & 0.635 & 0.648 & 0.594 & 0.728 & 0.735 & 0.779   \\
daf(Small+N)   & 0.561 & 0.578 & 0.505& 0.983 & 0.986 & 0.999 \\
daf(Small)   & 0.567 & 0.584 & 0.511& 0.800 & 0.809 & 0.846 \\
\hline
SDPED$_{7}$(Full+N) & 0.660 & 0.672 & 0.613 & 0.942 & 0.945 & 0.984   \\
SDPED$_{7}$(Full) & 0.666 & 0.679 & 0.634 & 0.722 & 0.734 & 0.765   \\
SDPED$_{7}$(Middle+N) & 0.662 & 0.677 & 0.638 & 0.933 & 0.937 & 0.975   \\
SDPED$_{7}$(Middle) & 0.659 & 0.673 & 0.633 & 0.709 & 0.724 & 0.777   \\
SDPED$_{7}$(Small+N)   & 0.628 & 0.642 & 0.596& 0.939 & 0.943 & 0.978 \\
SDPED$_{7}$(Small)   & 0.628 & 0.643 & 0.595& 0.808 & 0.817 & 0.902 \\
\hline
Avg(Full+N)   & 0.630 & 0.640 & 0.549& 0.914 & 0.918 & 0.960 \\
Avg(Full)   & 0.628 & 0.639 & 0.548& 0.738 & 0.745 & 0.786 \\
Avg(Middle+N)   & 0.622 & 0.634 & 0.557& 0.932 & 0.935 & 0.974 \\
Avg(Middle)   & 0.627 & 0.635 & 0.560& 0.763 & 0.772 & 0.825 \\
Avg(Small+N)   & 0.577 & 0.590 & 0.513& 0.952 & 0.955 & 0.985 \\
Avg(Small)   & 0.563 & 0.577 & 0.503& 0.746 & 0.754 & 0.803 \\
\hline
\end{tabular}
\end{table}

\begin{table}[htbp]\scriptsize
\renewcommand\arraystretch{1.25} 
\centering
\caption{\qquad \textbf{Results with error toleration 0.004 on UDED-$P_{1}$.} }
\label{UDED-aug}
\centering

\begin{tabular}{|p{32mm}<{\centering}|p{8mm}<{\centering}|p{8mm}<{\centering}|p{8mm}<{\centering}|p{8mm}<{\centering}|p{8mm}<{\centering}|p{8mm}<{\centering}|}
\hline
&\multicolumn{3}{|c|}{Test on Ordinary Data}&\multicolumn{3}{|c|}{Test on Noiseless Data}\\
\hline
Benchmarks   & ODS   & OIS   & AP  & ODS   & OIS   & AP \\
\hline
BDCN(Full+N) & 0.767 & 0.787 & 0.763 & 0.969 & 0.969 & 0.998 \\
BDCN(Full) & 0.775 & 0.797 & 0.776 & 0.943 & 0.946 & 0.977 \\
BDCN(Middle+N) & 0.739 & 0.773 & 0.742 & 0.971 & 0.972 & 0.998 \\
BDCN(Middle) & 0.757 & 0.770 & 0.742 & 0.933 & 0.934 & 0.996\\
BDCN(Small+N) & 0.645 & 0.696 & 0.655 & 0.967 & 0.968 & 0.999 \\
BDCN(Small) & 0.628 & 0.647 & 0.570 & 0.808 & 0.809 & 0.991 \\
\hline
Dexi(Full+N) & 0.771 & 0.784 & 0.766 & 0.955 & 0.962 & 0.995 \\
Dexi(Full) & 0.751 & 0.764 & 0.796 & 0.842 & 0.860 & 0.935 \\
Dexi(Middle+N) & 0.730 & 0.741 & 0.763 & 0.941 & 0.949 & 0.989 \\
Dexi(Middle) & 0.700 & 0.708 & 0.760 & 0.792 & 0.813 & 0.896 \\
Dexi(Small+N) & 0.712 & 0.723 & 0.776 & 0.847 & 0.922 & 0.937 \\
Dexi(Small) & 0.712 & 0.722 & 0.767 & 0.762 & 0.791 & 0.831 \\
\hline
EdgeNat(Full+N) & 0.709 & 0.714 & 0.601 & 0.831 & 0.831 & 0.996 \\
EdgeNat(Full) & 0.689 & 0.700 & 0.588 & 0.840 & 0.845 & 0.981 \\
EdgeNat(Middle+N) & 0.690 & 0.700 & 0.594 & 0.829 & 0.829 & 0.996 \\
EdgeNat(Middle) & 0.696 & 0.705 & 0.603 & 0.842 & 0.847 & 0.974 \\
EdgeNat(Small+N) & 0.675 & 0.688 & 0.580 & 0.824 & 0.826 & 0.996 \\
EdgeNat(Small) & 0.672 & 0.680 & 0.571 & 0.815 & 0.819 & 0.960 \\
\hline
daf(Full+N) & 0.750 & 0.776 & 0.719 & 0.971 & 0.972 & 0.998 \\
daf(Full) & 0.729 & 0.749 & 0.694 & 0.903 & 0.904 & 0.997 \\
daf(Middle+N) & 0.722 & 0.739 & 0.695 & 0.971 & 0.972 & 0.998 \\
daf(Middle) & 0.714 & 0.729 & 0.639 & 0.884 & 0.884 & 0.997 \\
daf(Small+N) & 0.586 & 0.630 & 0.512 & 0.920 & 0.920 & 1.000 \\
daf(Small) & 0.519 & 0.524 & 0.366 & 0.591 & 0.613 & 0.775\\
\hline
SDPED$_{7}$(Full+N) & 0.812 & 0.828 & 0.792 & 0.970 & 0.971 & 0.998\\
SDPED$_{7}$(Full) & 0.812 & 0.830 & 0.802 & 0.943 & 0.950 & 0.995\\
SDPED$_{7}$(Middle+N) & 0.804 & 0.820 & 0.798 & 0.969 & 0.970 & 0.998 \\
SDPED$_{7}$(Middle) & 0.794 & 0.806 & 0.794 & 0.940 & 0.942 & 0.994\\
SDPED$_{7}$(Small+N) & 0.758 & 0.782 & 0.721 & 0.972 & 0.973 & 0.998\\
SDPED$_{7}$(Small) & 0.758 & 0.770 & 0.722 & 0.929 & 0.934 & 0.992\\
\hline
Avg(Full+N) & 0.762 & 0.778 & 0.728 & 0.939 & 0.941 & 0.997 \\
Avg(Full) & 0.751 & 0.768 & 0.731 & 0.894 & 0.901 & 0.977\\
Avg(Middle+N) & 0.737 & 0.755 & 0.718 & 0.882 & 0.938 & 0.996 \\
Avg(Middle) & 0.732 & 0.744 & 0.708 & 0.878 & 0.884 & 0.971\\
Avg(Small+N) & 0.675 & 0.704 & 0.649 & 0.906 & 0.922 & 0.806\\
Avg(Small) & 0.658 & 0.669 & 0.599 & 0.781 & 0.793 & 0.910\\
\hline
\end{tabular}
\end{table}

\begin{table}[htbp]\scriptsize
\renewcommand\arraystretch{1.25} 
\centering
\caption{\qquad \textbf{Results with error toleration 0.001 on MDBD-$P_{1}$.} }
\label{MDBD-aug}
\centering

\begin{tabular}{|p{32mm}<{\centering}|p{8mm}<{\centering}|p{8mm}<{\centering}|p{8mm}<{\centering}|p{8mm}<{\centering}|p{8mm}<{\centering}|p{8mm}<{\centering}|}
\hline
&\multicolumn{3}{|c|}{Test on Ordinary Data}&\multicolumn{3}{|c|}{Test on Noiseless Data}\\
\hline
Benchmarks   & ODS   & OIS   & AP  & ODS   & OIS   & AP \\
\hline
BDCN(Full+N) & 0.277 & 0.278 & 0.104 & 0.832 & 0.832 & 1.000\\
BDCN(Full) & 0.278 & 0.280 & 0.106 & 0.612 & 0.613 & 0.771\\
BDCN(Middle+N) & 0.272 & 0.274 & 0.104 & 0.832 & 0.832 & 1.000\\
BDCN(Middle) & 0.287 & 0.290 & 0.123 & 0.547 & 0.550 & 0.632\\
BDCN(Small+N) & 0.277 & 0.280 & 0.118 & 0.832 & 0.832 & 1.000\\
BDCN(Small) & 0.258 & 0.261 & 0.099 & 0.497 & 0.498 & 0.643\\
\hline
Dexi(Full+N) & 0.305 & 0.307 & 0.126 & 0.829 & 0.829 & 1.000\\
Dexi(Full) & 0.299 & 0.300 & 0.120 & 0.609 & 0.610 & 0.861\\
Dexi(Middle+N) & 0.302 & 0.305 & 0.128 & 0.830 & 0.830 & 1.000\\
Dexi(Middle) & 0.303 & 0.305 & 0.128 & 0.635 & 0.635 & 0.799\\
Dexi(Small+N) & 0.340 & 0.342 & 0.185 & 0.856 & 0.856 & 0.994\\
Dexi(Small) & 0.335 & 0.337 & 0.204 & 0.558 & 0.560 & 0.626\\
\hline
EdgeNat(Full+N) & 0.277 & 0.279 & 0.101 & 0.815 & 0.815 & 0.993\\
EdgeNat(Full) & 0.274 & 0.275 & 0.098 & 0.634 & 0.635 & 0.845\\
EdgeNat(Middle+N) & 0.293 & 0.298 & 0.125 & 0.814 & 0.814 & 0.994\\
EdgeNat(Middle) & 0.296 & 0.299 & 0.119 & 0.718 & 0.719 & 0.933\\
EdgeNat(Small+N) & 0.298 & 0.300 & 0.128 & 0.813 & 0.814 & 0.993\\
EdgeNat(Small) & 0.286 & 0.288 & 0.117 & 0.648 & 0.649 & 0.873\\
\hline
daf(Full+N) & 0.293 & 0.296 & 0.116 & 0.832 & 0.832 & 1.000\\
daf(Full) & 0.282 & 0.284 & 0.103 & 0.604 & 0.606 & 0.750\\
daf(Middle+N) & 0.290 & 0.293 & 0.120 & 0.832 & 0.832 & 1.000\\
daf(Middle) & 0.298 & 0.300 & 0.139 & 0.720 & 0.723 & 0.892\\
daf(Small+N) & 0.316 & 0.321 & 0.163 & 0.832 & 0.832 & 0.979\\
daf(Small) & 0.287 & 0.290 & 0.135 & 0.755 & 0.763 & 0.982\\
\hline
SDPED$_{7}$(Full+N) & 0.327 & 0.330 & 0.165 & 0.832 & 0.832 & 1.000\\
SDPED$_{7}$(Full) & 0.330 & 0.333 & 0.170 & 0.788 & 0.792 & 0.957\\
SDPED$_{7}$(Middle+N) & 0.325 & 0.328 & 0.168 & 0.832 & 0.832 & 1.000\\
SDPED$_{7}$(Middle) & 0.322 & 0.325 & 0.163 & 0.806 & 0.808 & 0.963\\
SDPED$_{7}$(Small+N) & 0.315 & 0.318 & 0.155 & 0.832 & 0.832 & 1.000\\
SDPED$_{7}$(Small) & 0.315 & 0.318 & 0.158 & 0.778 & 0.780 & 0.969\\
\hline
Avg(Full+N) & 0.296 & 0.298 & 0.122 & 0.828 & 0.828 & 0.999\\
Avg(Full) & 0.293 & 0.294 & 0.119 & 0.649 & 0.651 & 0.837\\
Avg(Middle+N) & 0.296 & 0.300 & 0.129 & 0.828 & 0.828 & 0.999\\
Avg(Middle) & 0.301 & 0.300 & 0.134 & 0.685 & 0.687 & 0.844\\
Avg(Small+N) & 0.309 & 0.312 & 0.150 & 0.833 & 0.833 & 0.993\\
Avg(Small) & 0.296 & 0.299 & 0.143 & 0.647 & 0.650 & 0.819\\
\hline
\end{tabular}
\end{table}

\begin{table}[htbp]\scriptsize
\renewcommand\arraystretch{1.25} 
\centering
\caption{\qquad \textbf{Results with error toleration 0.001 on BIPED2-$P_{1}$.} }
\label{BIPED2-aug}
\centering

\begin{tabular}{|p{32mm}<{\centering}|p{8mm}<{\centering}|p{8mm}<{\centering}|p{8mm}<{\centering}|p{8mm}<{\centering}|p{8mm}<{\centering}|p{8mm}<{\centering}|}
\hline
&\multicolumn{3}{|c|}{Test on Ordinary Data}&\multicolumn{3}{|c|}{Test on Noiseless Data}\\
\hline
Benchmarks   & ODS   & OIS   & AP  & ODS   & OIS   & AP \\
\hline
BDCN(Full+N) & 0.654 & 0.657 & 0.519 & 0.983 & 0.983 & 0.998\\
BDCN(Full) & 0.644 & 0.648 & 0.503 & 0.955 & 0.957 & 0.992\\
BDCN(Middle+N) & 0.641 & 0.644& 0.501 & 0.983 & 0.983 & 0.998\\
BDCN(Middle) & 0.630 & 0.634 & 0.483 & 0.951 & 0.953 & 0.986\\
BDCN(Small+N) & 0.663 & 0.669 & 0.570 & 0.983 & 0.984 & 0.998\\
BDCN(Small) & 0.660 & 0.666 & 0.556 & 0.938 & 0.940 & 0.986\\
\hline
Dexi(Full+N) & 0.674 & 0.675 & 0.530 & 0.980 & 0.981 & 0.998\\
Dexi(Full) & 0.670 & 0.672 & 0.538 & 0.883 & 0.884 & 0.973\\
Dexi(Middle+N) & 0.668 & 0.670 & 0.523 & 0.977 & 0.977 & 0.999\\
Dexi(Middle) & 0.664 & 0.665 & 0.547 & 0.879 & 0.880 & 0.962\\
Dexi(Small+N) & 0.668 & 0.670 & 0.530 & 0.982 & 0.983 & 0.999\\
Dexi(Small) & 0.648 & 0.650 & 0.538 & 0.722 & 0.724 & 0.758 \\
\hline
EdgeNat(Full+N) & 0.590 & 0.592 & 0.423 & 0.891 & 0.892 & 0.965\\
EdgeNat(Full) & 0.589 & 0.592 & 0.388 & 0.830 & 0.831 & 0.962\\
EdgeNat(Middle+N) & 0.599 & 0.603 & 0.468 & 0.891 & 0.892 & 0.971\\
EdgeNat(Middle) & 0.604 & 0.607 & 0.468 & 0.836 & 0.831 & 0.955\\
EdgeNat(Small+N) & 0.600 & 0.604 & 0.471 & 0.890 & 0.891 & 0.970\\
EdgeNat(Small) & 0.600 & 0.603 & 0.475 & 0.825 & 0.826 & 0.940 \\
\hline
daf(Full+N) & 0.675 & 0.676 & 0.533 & 0.983 & 0.983 & 0.999\\
daf(Full) & 0.675 & 0.676 & 0.551 & 0.929 & 0.929 & 0.970\\
daf(Middle+N) & 0.673 & 0.675 & 0.557 & 0.983 & 0.983 & 0.998\\
daf(Middle) & 0.657 & 0.658 & 0.538 & 0.915 & 0.916 & 0.966\\
daf(Small+N) & 0.660 & 0.664 & 0.589 & 0.983 & 0.983 & 0.998\\
daf(Small) & 0.656 & 0.662 & 0.601 & 0.808 & 0.809 & 0.802\\
\hline
SDPED$_{7}$(Full+N) & 0.675 & 0.678 & 0.599 & 0.983 & 0.983 & 0.998\\
SDPED$_{7}$(Full) & 0.676 & 0.680 & 0.607 & 0.918 & 0.920 & 0.979\\
SDPED$_{7}$(Middle+N) & 0.669 & 0.674 & 0.604 & 0.983 & 0.983 & 0.998\\
SDPED$_{7}$(Middle) & 0.665 & 0.669 & 0.596 & 0.877 & 0.888 & 0.957\\
SDPED$_{7}$(Small+N) & 0.698 & 0.704 & 0.669 & 0.983 & 0.983 & 0.998\\
SDPED$_{7}$(Small) & 0.694 & 0.701 & 0.666 & 0.716 & 0.718 & 0.501\\
\hline
Avg(Full+N) & 0.654 & 0.656 & 0.521 & 0.964 & 0.964 & 0.992\\
Avg(Full) & 0.651 & 0.654 & 0.517 & 0.903 & 0.904 & 0.975\\
Avg(Middle+N) & 0.650 & 0.653 & 0.531 & 0.963 & 0.964 & 0.993\\
Avg(Middle) & 0.644 & 0.647 & 0.526 & 0.892 & 0.894 & 0.965\\
Avg(Small+N) & 0.658 & 0.662 & 0.566 & 0.964 & 0.965 & 0.993\\
Avg(Small) & 0.652 & 0.656 & 0.567 & 0.802 & 0.804 & 0.797\\
\hline
\end{tabular}
\end{table}

\subsubsection{Justification of the strict evaluation criterion}

Here, we justify the use of the strict error tolerance criterion when comparing the performance of models by showing that a stricter error tolerance distance can better distinguish model performance.

Tables \ref{BRIND-Maxdis} to \ref{BIPED2-Maxdis} exhibit the experimental results on BRIND, MDBD, and BIPED2, under various error tolerance settings. Average (Avg), Variant(Var), and normalized Variant (Var/Avg) are calculated across models, where higher Var/Avg represents better distinguishability of models. Partition 3 of the datasets is used because it is the most standard (8:2 of training and test images). Results show that reducing the error tolerance distance in evaluation allows better model distinguishability, and 1-pixel error tolerance should be the best.

\begin{table}[htbp]\scriptsize
\renewcommand\arraystretch{1.25} 
\centering
\caption{\qquad \textbf{Results on BRIND-$P_{3}$-400-100-E100, with different error toleration:} The columns correspond to error tolerance 1, 2, 3, 4, and 5 pixels on the diagonal, respectively. Results demonstrate that the ability to distinguish model performance increases as the error tolerance distance decreases.}
\label{BRIND-Maxdis}
\centering

\begin{tabular}{|p{28mm}<{\centering}|p{11mm}<{\centering}|p{11mm}<{\centering}|p{11mm}<{\centering}|p{11mm}<{\centering}|p{11mm}<{\centering}|}
\hline
\textit{maxDist}   & 0.003 & 0.005 & 0.0075& 0.010 & 0.0125 \\
\hline
\multicolumn{6}{|c|}{ODS}\\
\hline
HED & 0.663&0.742&0.781&0.798&0.816  \\
RCF & 0.665&0.745&0.784&0.800&0.819 \\
BDCN  & 0.663&0.736&0.773&0.791&0.810 \\
Dexi  & 0.663&0.750&0.789&0.806&0.823 \\
EdgeNat  & 0.597&0.720&0.772&0.791&0.810 \\
daf  & 0.655&0.727&0.764&0.782&0.801 \\
SDPED$\_7$  & 0.687&0.758&0.792&0.808&0.825 \\
Avg &0.656&0.740&0.779&0.797&0.815 \\
Var\ ($\times 10^{-3}$) &0.667&0.148&0.087&0.073&0.061 \\
Var/Avg\ ($\times 10^{-3}$) &1.017&0.200&0.111&0.091&0.073 \\
\hline
\multicolumn{6}{|c|}{OIS}\\
\hline
HED & 0.675& 0.756& 0.796& 0.815& 0.835   \\
RCF & 0.674& 0.757& 0.798& 0.816& 0.837  \\
BDCN  & 0.673& 0.749& 0.787& 0.806& 0.827  \\
Dexi  & 0.674& 0.763& 0.804& 0.822& 0.842  \\
EdgeNat  & 0.606& 0.733& 0.788& 0.808& 0.829  \\
daf  & 0.665& 0.738& 0.777& 0.796& 0.827  \\
SDPED$\_7$  & 0.698& 0.770& 0.805& 0.822& 0.841  \\
Avg &0.666&0.752&0.792&0.812&0.834 \\
Var ($\times 10^{-3}$) &0.697&0.150&0.091&0.076&0.035 \\
Var/Avg ($\times 10^{-3}$) &1.046&0.200&0.115&0.094&0.042 \\
\hline
\multicolumn{6}{|c|}{AP}\\
\hline
HED & 0.605& 0.741& 0.808& 0.829& 0.858   \\
RCF & 0.603& 0.739& 0.799& 0.827& 0.849  \\
BDCN  & 0.587& 0.713& 0.779& 0.802& 0.833  \\
Dexi  & 0.585& 0.723& 0.786& 0.814& 0.842  \\
EdgeNat  & 0.473& 0.662& 0.742& 0.776& 0.806  \\
daf  & 0.582& 0.710& 0.769& 0.792& 0.824  \\
SDPED$\_7$& 0.658& 0.781& 0.832& 0.853& 0.875 \\
Avg &0.585&0.724&0.788&0.813&0.841 \\
Var ($\times 10^{-3}$) &2.658&1.132&0.717&0.569&0.441 \\
Var/Avg ($\times 10^{-3}$) &4.547&1.563&0.910&0.700&0.525 \\
\hline
\end{tabular}
\end{table}

\begin{table}[htbp]\scriptsize
\renewcommand\arraystretch{1.25} 
\centering
\caption{\qquad \textbf{Results on MDBD-$P_{3}$-80-20-E100, with different error toleration:} The columns correspond to error toleration 1, 3, 5, 7, 9, and 11 pixels in diagonal, respectively. Results demonstrate that the ability to distinguish model performance increases as the error toleration distance decreases.}
\label{MDBD-Maxdis}
\centering

\begin{tabular}{|p{28mm}<{\centering}|p{11mm}<{\centering}|p{11mm}<{\centering}|p{11mm}<{\centering}|p{11mm}<{\centering}|p{11mm}<{\centering}|p{11mm}<{\centering}|}
\hline
\textit{maxDist}   & 0.001 & 0.003 & 0.005 & 0.007 & 0.009 & 0.011 \\
\hline
\multicolumn{7}{|c|}{ODS}\\
\hline
HED & 0.261&0.515&0.630&0.681&0.713& 0.735 \\
RCF & 0.245&0.492&0.606&0.657&0.689& 0.710  \\
BDCN  & 0.255&0.514&0.637&0.692&0.725& 0.748 \\
Dexi & 0.271&0.521&0.630&0.679&0.710& 0.732 \\
EdgeNat & 0.243&0.490&0.603&0.659&0.696& 0.724 \\
daf& 0.245&0.489&0.608&0.667&0.704& 0.730 \\
SDPED$\_7$& 0.291&0.543&0.644&0.687&0.716&0.736  \\
Avg &0.259&0.509&0.623&0.675&0.706&0.730 \\
Var ($\times 10^{-3}$) &0.262&0.344&0.235&0.161&0.129&0.117 \\
Var/Avg ($\times 10^{-3}$) &1.013&0.676&0.378&0.239&0.182&0.161 \\
\hline
\multicolumn{7}{|c|}{OIS}\\
\hline
HED & 0.264&0.520&0.637&0.689&0.722& 0.746 \\
RCF & 0.246&0.495&0.608&0.658&0.690&0.711  \\
BDCN & 0.257&0.515&0.638&0.693&0.728& 0.751 \\
Dexi & 0.274&0.529&0.640&0.690&0.722&0.745  \\
EdgeNat & 0.245&0.493&0.609&0.663&0.700& 0.728 \\
daf & 0.250&0.495&0.613&0.671&0.709& 0.736 \\
SDPED$\_7$& 0.297&0.561&0.669&0.717&0.751& 0.776 \\
Avg &0.262&0.515&0.631&0.683&0.717&0.742 \\
Var ($\times 10^{-3}$) &0.298&0.536&0.421&0.359&0.344&0.351 \\
Var/Avg ($\times 10^{-3}$) &1.138&1.040&0.667&0.525&0.479&0.473 \\
\hline
\multicolumn{7}{|c|}{AP}\\
\hline
HED& 0.094&0.356&0.507&0.575&0.625&0.655  \\
RCF & 0.079&0.307&0.444&0.501&0.540& 0.568 \\
BDCN & 0.088&0.339&0.493&0.571&0.615& 0.647 \\
Dexi& 0.102&0.374&0.518&0.582&0.623& 0.652 \\
EdgeNat& 0.082&0.314&0.452&0.527&0.571& 0.611 \\
daf & 0.084&0.316&0.462&0.538&0.591&0.625  \\
SDPED$\_7$& 0.139&0.477&0.639&0.701&0.740& 0.767 \\
Avg &0.095&0.355&0.502&0.571&0.627&0.619 \\
Var ($\times 10^{-3}$) &0.369&3.001&3.799&3.566&4.063&3.584 \\
Var/Avg ($\times 10^{-3}$) &3.862&8.461&7.566&6.248&6.476&5.792 \\
\hline
\end{tabular}
\end{table}

\begin{table}[htbp]\scriptsize
\renewcommand\arraystretch{1.25} 
\centering
\caption{\qquad \textbf{Results on BIPED2-$P_{3}$-200-50-E75, with different error toleration:} The columns correspond to error toleration 1, 3, 5, 7, and 9 pixels in diagonal, respectively. Results demonstrate that the ability to distinguish model performance increases as the error toleration distance decreases.}
\label{BIPED2-Maxdis}
\centering

\begin{tabular}{|p{28mm}<{\centering}|p{11mm}<{\centering}|p{11mm}<{\centering}|p{11mm}<{\centering}|p{11mm}<{\centering}|p{11mm}<{\centering}|}
\hline
\textit{maxDist}   & 0.001 & 0.003 & 0.005& 0.007 & 0.009 \\
\hline
\multicolumn{6}{|c|}{ODS}\\
\hline
HED & 0.640&0.830&0.864&0.881&0.893   \\
RCF& 0.640&0.832&0.865&0.882&0.894  \\
BDCN & 0.655&0.826&0.859&0.877&0.890 \\
Dexi  & 0.677&0.839&0.869&0.886&0.899 \\
EdgeNat & 0.575&0.796&0.833&0.852&0.866  \\
daf & 0.663&0.821&0.856&0.875&0.889  \\
SDPED$\_7$ & 0.673&0.831&0.862&0.879&0.892 \\
Avg &0.646&0.825&0.858&0.876&0.889 \\
Var ($\times 10^{-3}$) &1.025&0.166&0.122&0.107&0.097 \\
Var/Avg ($\times 10^{-3}$) &1.586&0.202&0.142&0.122&0.109 \\
\hline
\multicolumn{6}{|c|}{OIS}\\
\hline
HED & 0.644&0.837&0.872&0.892&0.900  \\
RCF & 0.644&0.838&0.874&0.892&0.905  \\
BDCN & 0.659&0.831&0.867&0.886&0.901  \\
Dexi & 0.679&0.842&0.875&0.892&0.906  \\
EdgeNat& 0.579&0.802&0.840&0.861&0.876   \\
daf& 0.666&0.827&0.863&0.884&0.899  \\
SDPED$\_7$& 0.680&0.840&0.872&0.891&0.905  \\
Avg &0.650&0.831&0.866&0.885&0.899 \\
Var ($\times 10^{-3}$) &1.027&0.163&0.129&0.109&0.094 \\
Var/Avg ($\times 10^{-3}$) &1.580&0.197&0.149&0.123&0.104 \\
\hline
\multicolumn{6}{|c|}{AP}\\
\hline
HED& 0.494&0.816&0.866&0.895&0.911  \\
RCF& 0.497&0.814&0.868&0.895&0.910  \\
BDCN& 0.512&0.814&0.868&0.891&0.909  \\
Dexi& 0.530&0.769&0.826&0.848&0.864  \\
EdgeNat& 0.408&0.724&0.783&0.817&0.844  \\
daf & 0.535&0.793&0.849&0.882&0.893 \\
SDPED$\_7$& 0.588&0.853&0.897&0.920&0.930  \\
Avg &0.509&0.796&0.851&0.878&0.894 \\
Var ($\times 10^{-3}$) &2.562&1.457&1.167&1.021&0.781 \\
Var/Avg ($\times 10^{-3}$) &5.032&1.827&1.372&1.163&0.873 \\
\hline
\end{tabular}
\end{table}

\section{Discussion}

In this section, we discuss plausible factors contributing to the effectiveness of the SDPED architecture, analyze its parameter efficiency and robustness across diverse scenarios, and identify limitations for future research.

\subsection{Factors Contributing to the Effectiveness of SDPED}

The observed performance improvements of SDPED, particularly in the AP metric, can be attributed to four core design philosophies:

\subsubsection{Avoiding Information Loss by Removing Down-sampling}

Unlike most previous ED models that rely on down-sampling (e.g., max-pooling) to extract multi-scale features, SDPED eliminates these operations entirely to maintain full-resolution processing throughout the pipeline. Down-sampling inherently introduces spatial uncertainty; for instance, a $2\times 2$ max-pooling operation collapses four pixels into one, which results in information loss and makes the precise recovery of fine edge locations highly challenging. While down-sampling was traditionally justified for expanding the receptive field \cite{R2008Multi}, deep CNN architectures already possess a sufficiently large receptive field to capture multi-scale context. By removing down-sampling, SDPED preserves the original spatial information, thereby maximizing the localization precision required for high-frequency edge features.

\subsubsection{Task-Adaptive Texture Differentiation}

The primary challenge in ED is distinguishing between edges and textures. By performing a task-adaptive architectural transfer from image SR tasks, SDPED leverages the CSDB backbone modified from the DB, which has been extensively validated for its high-frequency feature handling capabilities. The logic behind this transfer is that both SR and ED share a fundamental sub-task: the precise localization of texture-rich regions. In SR, these regions are identified for enhancement, while in ED, they are identified for suppression. Our results suggest that an architecture capable of understanding texture for reconstruction can also be effectively adapted for texture suppression in detection tasks. Specifically, a model that excels in SR must accurately localize high-frequency textures to restore details; the same localization capability can be leveraged to differentiate these textures from edges.

\subsubsection{Consistency-based Response Filtering}

Essentially, the distinction between edges and textures is based on response intensity to specific criteria, such as gradients, but textural regions can also produce high responses, leading to false positives. Here, the CSDB structure acts as an implicit consistency-filtering mechanism. Because it cascades multiple SDB blocks with skipping connections, the final prediction effectively reinforces pixels that are consistently identified as edges across multiple processing stages. If a textural pixel produces a high gradient response in only a single layer but lacks consistency in subsequent blocks, its final response is attenuated through the integration mechanism. This design encourages the preservation of more reliable edge responses, effectively filtering out sporadic textural noise that often plagues shallower or non-cascaded models.

\subsubsection{Restoring Structural Balance in Feature Fusion}

ED architectures can be broadly categorized into feature extraction and feature fusion stages. However, there is a significant structural imbalance in previous ED models: millions of parameters are often dedicated to feature extraction, yet the critical fusion stage is reduced to a single $1\times 1$ convolution layer. This bottleneck may lead to the underutilization of rich extracted features. By incorporating a more robust fusing block with three convolutional layers (including $3 \times 3$), SDPED restores the balance between extraction and fusion, allowing for a more sophisticated synthesis of edge features and resulting in sharper, more accurate final edge maps.

\subsection{Model Size and Computational Cost}

As summarized in Table \ref{Size}, SDPED exhibits strong parameter efficiency. It achieves SOTA performance with the fewest parameters among all compared models—less than 40\% of the second-smallest model (HED) and only 1.0\% of the largest (EdgeNat). This suggests that SDPED utilizes its parameters more efficiently than previous backbones, namely, with less redundancy to achieve high-precision edge localization. However, SDPED also displays higher computational cost. Even SDPED$_{3}$ has floating-point operations (FLOPs) about three times as many as RIND and nine times as many as HED. Therefore, it shows the effectiveness of compressing the model size but not the computational cost.

\begin{table}[htbp]\small
\renewcommand\arraystretch{1.25}
\centering
\caption{\qquad \textbf{Comparisons of model sizes and FLOPs:} SDPED$\_7$ has only 38.9$\%$ of the parameters of the second-smallest model (HED), yet significantly outperforms it. For more details, the size of SDPED$\_7$ is about 38.9$\%$ of RCF, 9.6$\%$ of RIND, 35.1$\%$ of BDCN, 16.3$\%$ of Dexi, 1.0$\%$ of EdgeNat, and 32.8\% of daf. However, SDPED also has relatively larger FLOPs, and even SDPED$_{3}$ needs about 3$\times$ FLOPs compared to the second-largest one (RIND).}
\label{Size}
\begin{tabular}{|p{32mm}<{\centering}|p{16mm}<{\centering}|p{40mm}<{\centering}|p{32mm}<{\centering}|}
\hline
\          & Size (kB) & Num. of Coe. ($\times 10^{6}$) & FLOPs (GFLOPs)\\
\hline
HED & 57499 & 14.716171 & 63 \\
\hline
RCF & 57850 & 14.803781 & 80 \\
\hline
RIND & 232593 & 59.388526 & 180  \\
\hline
BDCN & 63749 & 16.302120 & 112 \\
\hline
Dexi & 137775 & 35.215245 & 101 \\
\hline
EdgeNat & 2349905 & 587.34319 & 92  \\
\hline
daf     &68375  & 17.465908 & 116\\
\hline
SDPED$\_3$ (Ours) & 10403 & 2.651946 & 543 \\
\hline
SDPED$\_5$ (Ours) & 16436 & 4.190292 & 858 \\
\hline
SDPED$\_7$ (Ours) & 22470 & 5.728638 & 1173 \\
\hline
\end{tabular}
\end{table}

\subsection{Robustness and Cross-Dataset Adaptability}

One of the key strengths of SDPED is its robustness across multiple ED datasets. As highlighted in \cite{MK2022Have}, previous models often exhibit dataset-dependent performance, with varying behavior across benchmarks. For instance, HED and RCF achieve the highest accuracy on BRIND but are outperformed on UDED and BIPED2. Similarly, BDCN performs best on UDED but lags on other datasets, while Dexi and RIND yield superior results on MDBD and BIPED2 but fail to maintain consistency across different benchmarks.

In contrast, SDPED demonstrates consistently competitive, and often state-of-the-art, performance across various datasets. It yields the best results in all benchmarks on BRIND, UDED, and MDBD and significantly outperforms previous models in AP on BIPED2 while maintaining competitive performance on other benchmarks. This demonstrates its adaptability across different data distributions. These results indicate that SDPED is not only highly precise but also versatile, making it a robust solution for ED tasks across diverse datasets.

\subsection{Limitations and Future Work}

While our framework sets a new benchmark for high-precision ED, several avenues for future work remain:

\subsubsection{Inference Efficiency and Memory Footprint}

The primary trade-offs for SDPED’s high precision are its memory consumption and inference speed. Since the model operates on ordinary resolution of images without down-sampling, it requires more GPU memory and inference time compared to models that utilize pooling. While this study prioritized precision over memory and speed, future work should explore memory-efficient implementations to adapt SDPED for applications on resource-constrained devices.

\subsubsection{Advanced Fusion Mechanisms}

Although our multi-layer fusing block improves performance, as confirmed by ablation studies, it remains a relatively straightforward convolutional integration. Given that feature fusion could be the bottleneck in high-precision tasks, exploring more diverse and advanced schemes—such as attention-based fusion or dynamic feature selection—could further refine the ability of leveraging extracted rich edge information.

\subsubsection{The Data Quality Bottleneck}

A fundamental challenge in ED tasks is the lack of a formal definition for edges. If edges had a strict mathematical definition, traditional algorithms would have already solved ED problems. As a result, ED datasets heavily rely on human-labeled annotations, which inherently introduce noise. As demonstrated in Section \ref{sec:Experiment}, model performance is often correlated with dataset quality. For instance, models perform best on UDED, despite it having the fewest training samples, likely due to its superior annotation quality. This suggests that data quality may represent a bottleneck comparable to model architecture in ED tasks. While our ideal prior method in Section \ref{sec:Noiseless} provides a way to leverage noiseless data, the development of meticulously curated datasets and data-refinement approaches remains essential for further performance breakthroughs in ED.

 \section{Conclusion}

In this paper, we have presented a comprehensive framework for high-precision edge detection (ED), with extensive experiments:

(1) We introduce an ideal prior guidance strategy—a data-centric augmentation paradigm that incorporates noiseless ground-truth information into the training process. By treating labels as noise-free input samples, this approach provides a practical mechanism to bypass the human-annotation ceiling, which inevitably introduces subjectivity and noise into supervision. More broadly, this perspective suggests that ground-truth information can be exploited as an ideal physical prior to guide learning beyond the limitations imposed by noisy supervision, providing a potential solution for a wide range of learning tasks, particularly for those where human annotation is unavoidable.

(2) By performing a task-adaptive re-engineering of texture-handling structures originally developed for image super-resolution (SR), the proposed SDPED model, built upon the Cascaded Skipping Density Block (CSDB), is able to suppress textural responses while preserving high spatial fidelity, and can achieve strong performance, particularly in Average Precision (AP), while maintaining high parameter efficiency. It indicates that a precisely tailored and compact architecture can be more effective than substantially larger models, and that certain architectural backbones exhibit meaningful task transferability across low-level vision problems.

(3) We address inconsistencies in ED model assessment by analyzing various evaluation parameters. By shifting evaluation from resolution-dependent proportional ratios to a fixed-pixel standard, we offer a more rigorous and resolution-independent benchmark for high-precision ED. Moreover, by comparing model performance across various evaluation settings, we demonstrate that a stricter error tolerance distance can better distinguish the abilities of different models.

Taken together, this work provides a coherent study of high-precision edge detection from architectural, data-centric, and evaluation perspectives. We hope these insights contribute to the development of more reliable and precise edge detectors and inspire further research on precision-oriented modeling in low-level and soft computing–based vision tasks.

\balance

\section{Bibliographies}
\bibliographystyle{unsrt}
\bibliography{EDBitex}

\end{document}